\documentclass[10pt,twocolumn,letterpaper]{article}

\usepackage{iccv}
\usepackage{times}
\usepackage{epsfig}
\usepackage{graphicx}
\usepackage{amsmath}
\usepackage{amssymb}
\usepackage[utf8]{inputenc} 
\usepackage[T1]{fontenc}    
\usepackage{CJKutf8}        
\usepackage{booktabs}
\usepackage{url}            
\usepackage{booktabs}       
\usepackage{amsfonts}       
\usepackage{nicefrac}       
\usepackage{microtype}      
\usepackage{xcolor}
\usepackage{multirow}
\usepackage{tabularx}
\usepackage{adjustbox}
\usepackage{wrapfig}
\usepackage{caption}        

\usepackage[accsupp]{axessibility}

\usepackage{makecell}
\usepackage{pifont}
\usepackage[abs]{overpic}

\usepackage[pagebackref=true,breaklinks=true,letterpaper=true,colorlinks,bookmarks=false]{hyperref}

\iccvfinalcopy 

\def\httilde{\mbox{\tt\raisebox{-.5ex}{\symbol{126}}}}

\ificcvfinal\fi

\begin{document}

\title{Iterative Prompt Learning for Unsupervised Backlit Image Enhancement}

\author{Zhexin Liang \quad Chongyi Li \quad Shangchen Zhou \quad Ruicheng Feng \quad Chen Change Loy\\
S-Lab, Nanyang Technological University\\
{\tt\small \{zliang008, s200094, ruicheng002, ccloy\}@ntu.edu.sg, lichongyi25@gmail.com}\\
{\tt\small \url{https://zhexinliang.github.io/CLIP_LIT_page/}}
}

\twocolumn[{%
   \renewcommand\twocolumn[1][]{#1}%
   \maketitle
   \vspace{-10mm}
   \begin{center}
    \centering
    \includegraphics[width=\linewidth]{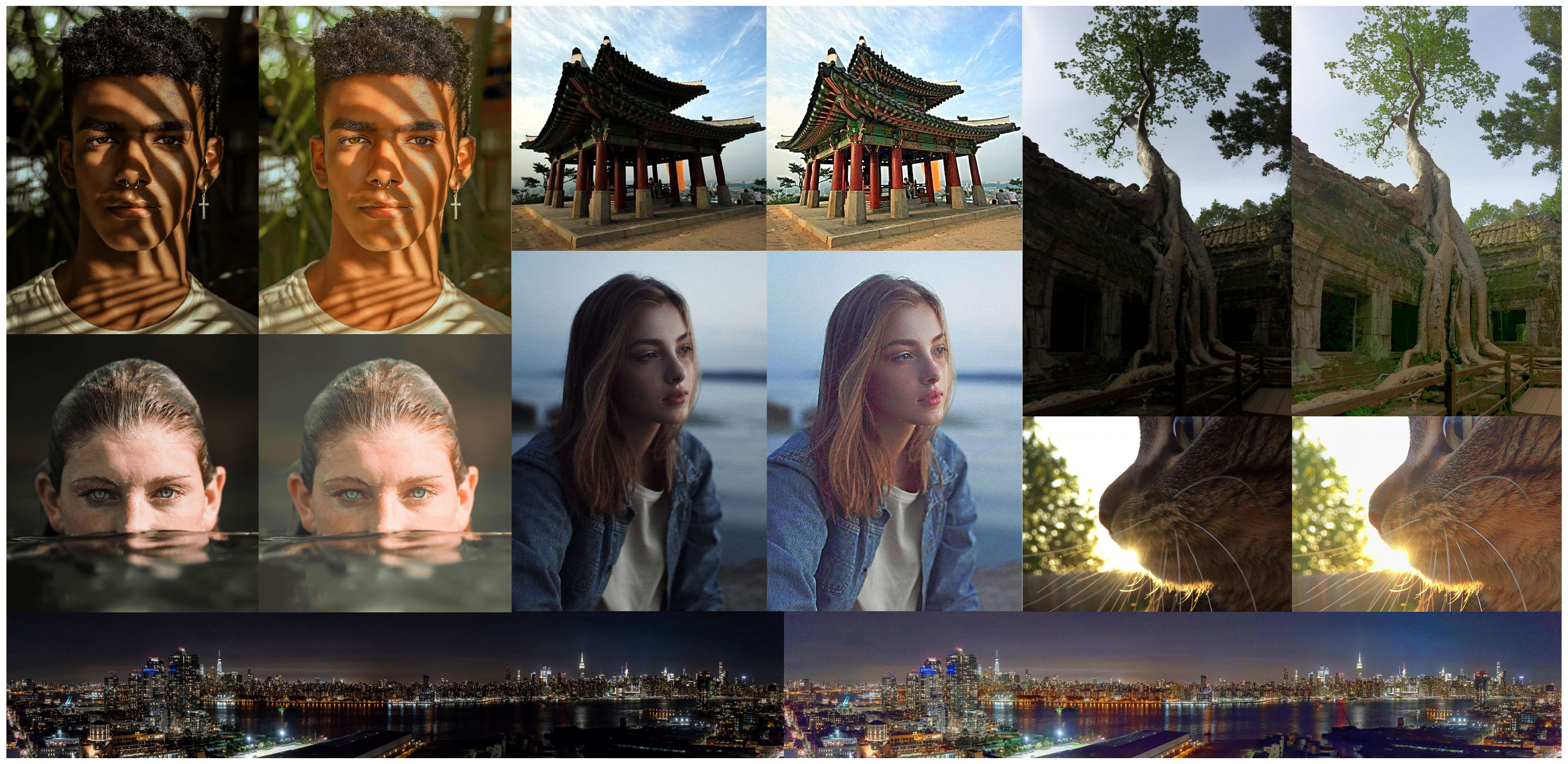}
    \vspace{-7mm}
    \captionof{figure}{The proposed method trained using only hundreds of images yields favorable results on unseen backlit images captured in various scenarios, including the human face, natural landscape, animal, architecture, and night scene.
 }
 \vspace{1mm}
    \label{fig:teaser}
   \end{center}%
}]
  %
\thispagestyle{empty} 
\begin{abstract}
\vspace{-2mm}
We propose a novel unsupervised backlit image enhancement method, abbreviated as CLIP-LIT, by exploring the potential of Contrastive Language-Image Pre-Training (CLIP) for pixel-level image enhancement.
We show that the open-world CLIP prior not only aids in distinguishing between backlit and well-lit images, but also in perceiving heterogeneous regions with different luminance, facilitating the optimization of the enhancement network.
Unlike high-level and image manipulation tasks, directly applying CLIP to enhancement tasks is non-trivial, owing to the difficulty in finding accurate prompts.
To solve this issue, we devise a prompt learning framework that first learns an initial prompt pair by constraining the text-image similarity between the prompt (negative/positive sample) and the corresponding image (backlit image/well-lit image) in the CLIP latent space.
Then, we train the enhancement network based on the text-image similarity between the enhanced result and the initial prompt pair.
To further improve the accuracy of the initial prompt pair, we iteratively fine-tune the prompt learning framework to reduce the distribution gaps between the backlit images, enhanced results, and well-lit images via rank learning, boosting the enhancement performance. 
Our method alternates between updating the prompt learning framework and enhancement network until visually pleasing results are achieved.
Extensive experiments demonstrate that our method outperforms state-of-the-art methods in terms of visual quality and generalization ability, without requiring any paired data. 
\end{abstract}

\vspace{-4mm}
\section{Introduction}
\label{sec:intro}

Backlit images are captured when the primary light source is behind some objects.
The images often suffer from highly imbalanced illuminance distribution, which affects the visual quality or accuracy of subsequent perception algorithms.

Correcting backlit images manually is a laborious task  given the intricate challenge of preserving the well-lit regions while enhancing underexposed regions.
One could apply an automatic light enhancement approach but will find that existing approaches could not cope well with backlit images \cite{survey}.
For instance, many existing supervised light enhancement methods \cite{Chen2018, URetinexNet,KinD} cannot precisely perceive the bright and dark areas, and thus process these regions using the same pipeline, causing 
over-enhancement in well-lit areas or under-enhancement in low-light areas.
Unsupervised light enhancement methods, on the other hand, either rely on ideal assumptions such as average luminance and a gray world model \cite{Guo2020CVPR, ZeroDCE++} or directly learn the distribution of reference images via adversarial training \cite{Jiang2019}.
The robustness and generalization capability of these methods are limited.
As for conventional exposure correction methods \cite{Mahoud2021,Lu2012}, they struggle in coping with real-world backlit images due to the diverse backlit scenes and luminance intensities.
The problem cannot be well resolved by collecting backlit images that consist of ground truth images that are retouched by photographers \cite{LV2022103403}, since these images can never match the true distribution of 
real backlit photos.

\begin{figure}[t]
    \vspace{-0.6em}
    \begin{center}
        \includegraphics[width=.95\linewidth]{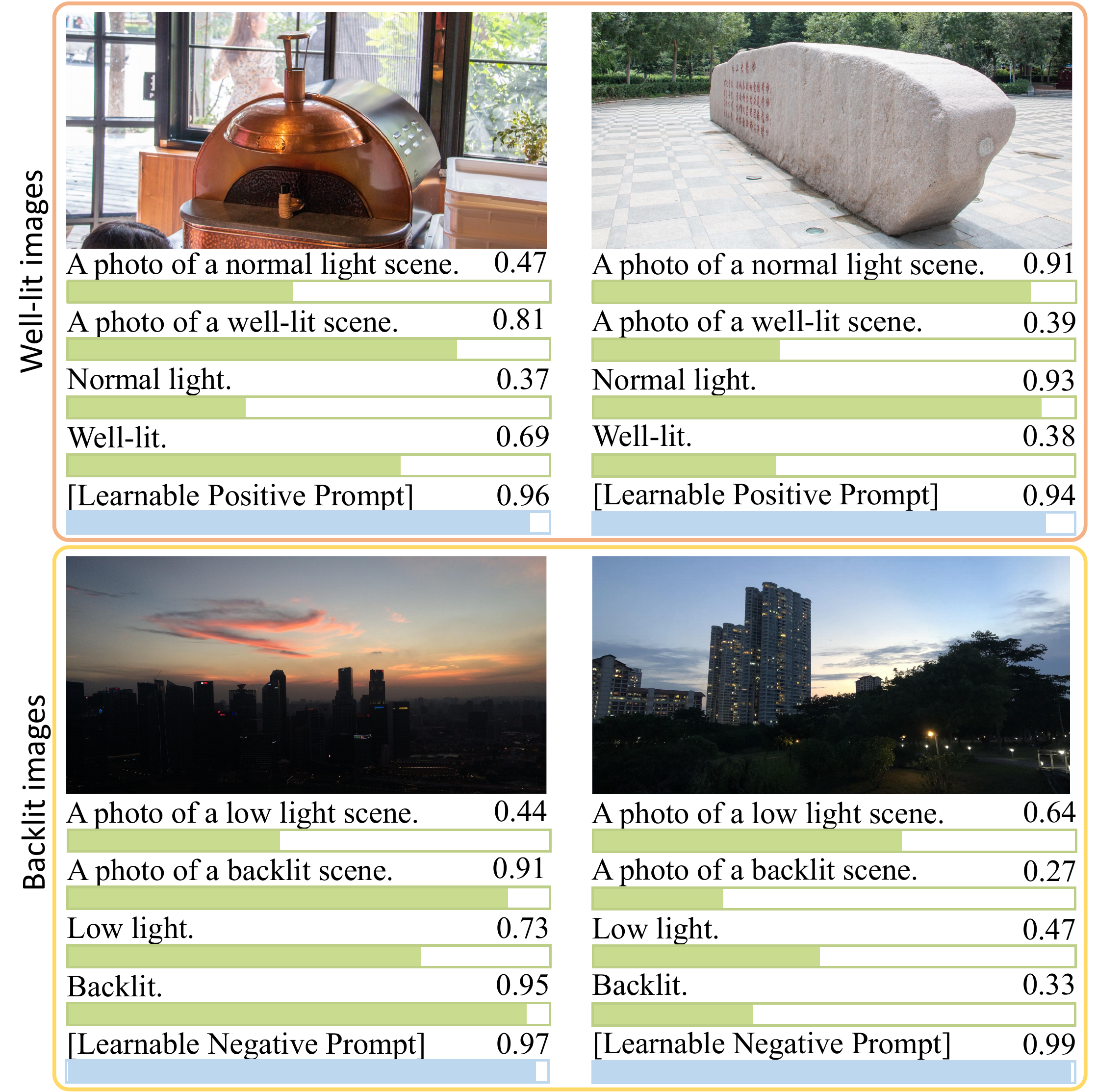}
    \end{center}
    \vspace{-.6cm}
    \caption{Motivation. CLIP scores of proper prompts demonstrate alignment with human annotations (\eg, well-lit images), suggesting that CLIP can serve as an indicator to differentiate between well-lit and backlit images. However, the best wordings could differ on a case-by-case basis due to complex illumination. In contrast, the learnable positive/negative prompts are more robust and consistent with the labels.}
    \label{fig:motivation}
    \vspace{-1.8em}
\end{figure}

In this work, we propose an unsupervised method for backlit image enhancement. 
Different from previous unsupervised methods that learn curves or functions based on some physical hypothesis or learn the distribution of well-lit images via adversarial training that relies on task-specific data, we explore the rich visual-language prior encapsulated in a Contrastive Language-Image Pre-Training (CLIP) \cite{CLIP} model for pixel-level image enhancement.
While CLIP can serve as an indicator to distinguish well-lit and backlit images to a certain extent, using it directly for training a backlit image enhancement network is still non-trivial.
For example, for a well-lit image (Fig.~\ref{fig:motivation} top left), replacing similar concepts ``normal light'' with ``well-lit'' brings a huge increase in CLIP score. In the opposite case (Fig.~\ref{fig:motivation} top right), ``normal light'' becomes the correct prompt.
This indicates the optimal prompts could vary on a case-by-case basis due to the complex illuminations in the scene.
In addition, it is barely possible to find accurate `word' prompts to describe the precise luminance conditions.
Prompt engineering is labor-intensive and time-consuming to annotate each image in the dataset.
Moreover, the CLIP embedding is often interfered by high-level semantic information in an image.
Thus, it is unlikely to achieve optimal performance with fixed prompts or prompt engineering.

To overcome the problems, we present a new pipeline to tailor the CLIP model for our task. It consists of the following components:
1) \textit{Prompt Initialization.} We first encode the backlit and well-lit images along with a learnable prompt pair (positive and negative samples) into the latent space using the pre-trained CLIP's image and text encoder.
By narrowing the distance between the images and text in the latent space, we obtain an initial prompt pair that can effectively distinguish between backlit and well-lit images.
2) \textit{CLIP-aware Enhancement Training.} With the initialized prompt, we train an enhancement network using the text-image similarity constraints in the CLIP embedding space.
3) \textit{Prompt Refinement.} We introduce a prompt fine-tuning mechanism, in which we update the prompt by further distinguishing the distribution gaps among backlit images, enhanced results, and well-lit images via rank learning.
We iteratively update the enhancement network and prompt learning framework until achieving visually pleasing results.

\begin{figure*}[t]
    \vspace{-1em}
    \centering
    \includegraphics[width=1\linewidth]{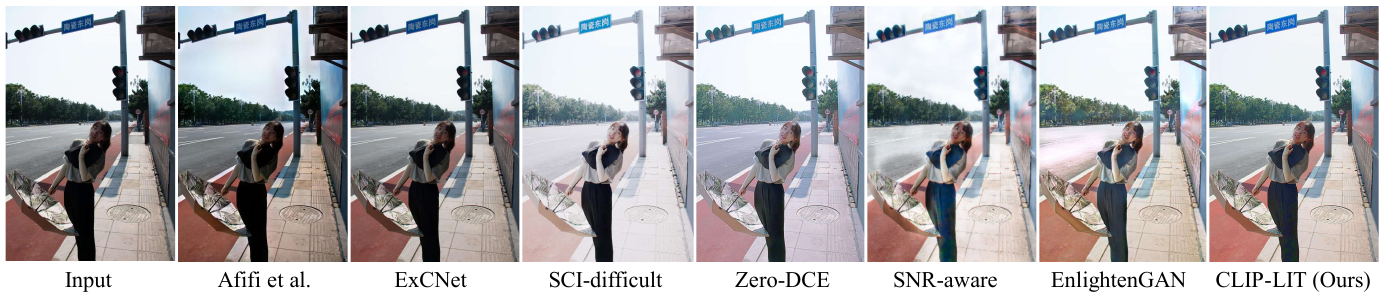}
    \vspace{-2em}
    \caption{\small{Visual comparison between our method and the state-of-the-art light enhancement methods, including exposure correction method (Afifi et al. \cite{Mahoud2021}), backlit enhancement method (ExCNet \cite{zhang2019zero}), low-light image enhancement methods (SCI \cite{SCI2022}, Zero-DCE \cite{Guo2020CVPR}, SNR-aware \cite{SNR2022}, EnlightenGAN \cite{Jiang2019}). Our method effectively enhances the backlit image without introducing artifacts and over-/under-enhancement.}}
    \vspace{-0.5em}
    \label{fig:intro}
\end{figure*}

\begin{figure*}[!t]
    \begin{center}
        \includegraphics[width=\linewidth]{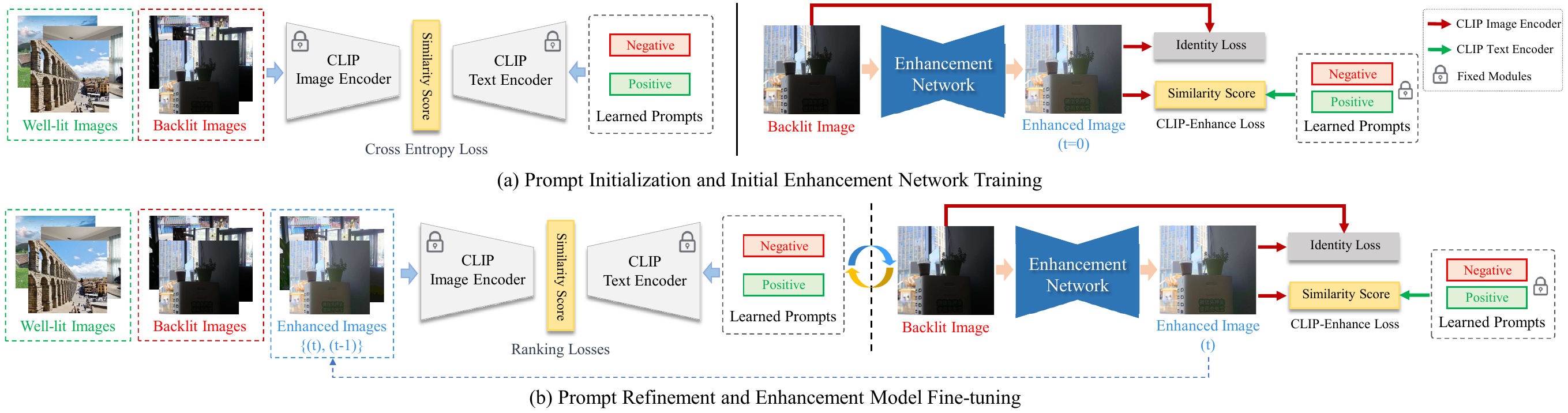}
    \end{center}
    \vspace{-1.8em}
    \caption{Our proposed method involves two main stages. (a) The first stage constitutes prompt initialization and the initial training of an enhancement network. (b) The second stage involves prompt refinement and enhancement model fine-tuning. The two components here are updated in an alternating manner. 
    The prompt refinement in the second stage aims at learning accurate prompts that distinguish among backlit images, enhanced results, and well-lit images.
    By employing these learned prompts, the enhancement network produces enhanced results that are similar to well-lit images and distinct from backlit images in the CLIP embedding space, ultimately leading to visually pleasing results. 
    }
    \label{fig:framework_structure}
    \vspace{-1.5em}
\end{figure*}

Our method stands apart from existing backlit image enhancement techniques as we leverage the intrinsic perceptual capability of CLIP. Rather than solely utilizing CLIP as a loss objective \cite{frans2021clipdraw,zhou2022maskclip}, we incorporate prompt refinement as an essential component of the optimization process to further enhance performance. Our method is the first work to utilize prompt learning and the CLIP prior into the low-level vision task. Our approach surpasses state-of-the-art methods in both qualitative and quantitative metrics, without requiring any paired training data.
We demonstrate the generalization capability and robustness of our method through the preview of our results shown in Fig. \ref{fig:teaser}, and we compare our results with existing methods in Fig. \ref{fig:intro}.

\vspace{-0.5em}
\section{Related Work}
\vspace{-0.5em}
\noindent
\textbf{Backlit Image Enhancement.} 
Several approaches have been proposed in the literature. Li and Wu \cite{Libacklit} employ a region segmentation technique in combination with a learning-based restoration network to separately process the backlit and front-lit areas of an image.
Buades et al. \cite{buades2020backlit} and Wang et al. \cite{wang2016fusion} use fusion-based techniques to combine pre-processed images.
Zhang et al. \cite{zhang2019zero} learn a parametric ``S-curve'' using a small image-specific network,  ExCNet, to correct ill-exposed images.
More recently, Lv et al. \cite{LV2022103403} have created the first paired backlit dataset, named BAID, in which the ground truth images are edited by photographers so that the quality is still sub-optimal, shown in Fig.~\ref{fig:comparison}.

\noindent
\textbf{Light Enhancement.} 
Backlit image enhancement is closely related to low-light image enhancement and exposure correction. Traditional methods for low-light image enhancement \cite{Fu2016,Li2018} typically employ the Retinex model to restore normal-light images. With the availability of paired data~\cite{li2023embedding,Chen2018} and simulated data~\cite{zhou2022lednet}, several supervised methods \cite{URetinexNet,SNR2022} have been proposed, which design various networks for low-light image enhancement.
Despite their success, supervised methods suffer from limited generalization capability. Consequently, unsupervised methods\cite{Guo2020CVPR, ZeroDCE++,RUAS2021, SCI2022} have garnered increasing attention.
Since low-light image enhancement cannot effectively process both underexposed and overexposed regions, exposure correction methods \cite{Mahoud2021,SICE,Lu2012} have also been proposed. For example, Afifi et al. \cite{Mahoud2021} propose an exposure correction network based on Laplacian pyramid decomposition and reconstruction.

\noindent
\textbf{CLIP and Prompting in Vision.} 
CLIP \cite{CLIP} has shown remarkable performance in zero-shot classification, thanks to the knowledge learned from large-scale image-text data. Its generalizability has been shown in high-level tasks\cite{zang2022open,kuoopen,zhou2022maskclip}. 
A recent study\cite{wang2022exploring} shows that the rich visual language prior encapsulated in CLIP can be used for assessing both the quality and abstract perception  of images in a zero-shot manner. These studies inspire our work to exploit CLIP for backlit image enhancement. 
Prompt learning, as the core of vision-and-language models, is a recent emerging research direction. CoOp \cite{CoOp} introduces prompt learning into the adaptation of vision-language models for downstream vision tasks. CoCoOp \cite{CoCoOp} further improves the generalizability by allowing a prompt to be conditioned on each input instance rather than fixed once learned. 
Existing prompt learning methods focus solely on obtaining better prompts for high-level vision tasks.
In contrast, our approach uses prompt learning to extract more accurate low-level image representations, such as color, exposure, and saturation, while ignoring high-level semantic information in CLIP.

\begin{figure*}[!t]
 \vspace{-1.2em}
    \begin{center}
          \includegraphics[width=1.0\linewidth]{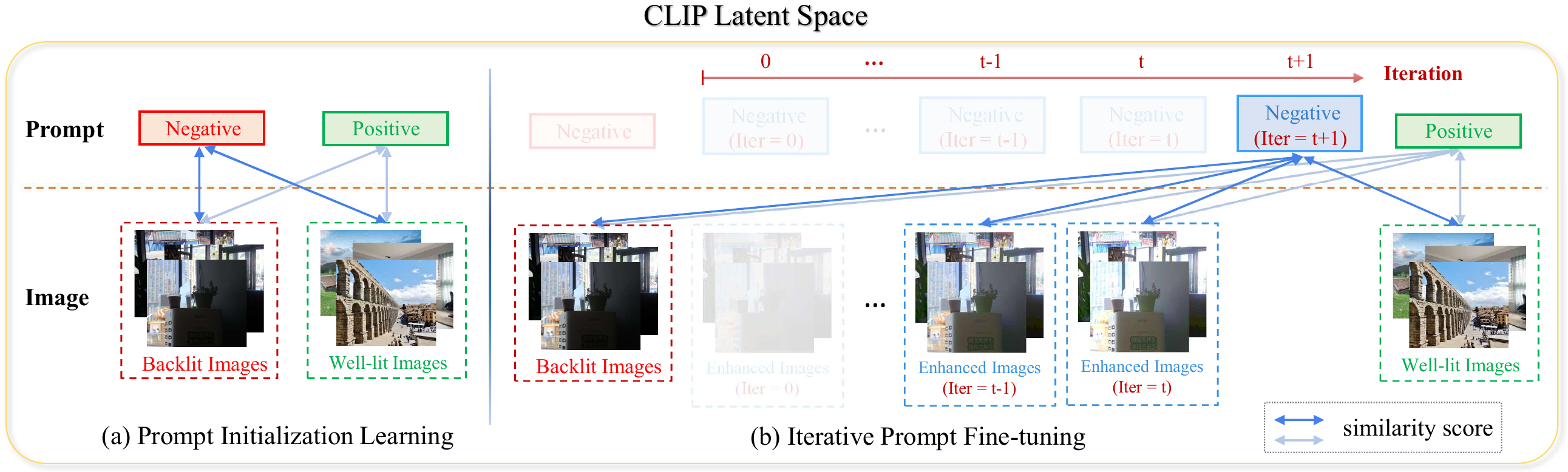}
    \end{center}
    \vspace{-1.5em}
    \caption{Illustration of the prompt learning framework. \textbf{1.} Prompt Initialization. A cross-entropy loss constrains the learned prompts, maximizing the distance between the representation of negative and positive samples in the CLIP latent space. \textbf{2.} Adding the enhanced results from the current round $I_{t}$ into the ranking process (i.e., ranking loss) to make the enhanced results $I_{t}$ closer to the representation of the well-lit images $I_w$ in the CLIP latent space and far from the representation of the input image $I_b$. \textbf{3.} Adding the images inferred from the previous round $I_{t-1}$ to constrain the result of updated enhancement network. $I_{t}$ being closer to the representation of positive samples than the previous round $I_{t-1}$, and far from the representation of negative ones $I_{b}$ in CLIP latent space. }
    \label{fig:iterative_prompt_learning_framework}
    \vspace{-1.8em}
\end{figure*}

\section{Methodology}
\noindent\textbf{Overview.} 
Our proposed approach consists of two stages, as illustrated in Fig.~\ref{fig:framework_structure}.
In the first stage, we learn an initial prompt pair (negative/positive prompts referring to backlit/well-lit images) by constraining the text-image similarity between the prompt and the corresponding image in the CLIP embedding space.
With the initial prompt pair, we use a frozen CLIP model to compute the text-image similarity between the prompts and the enhanced results to train the initial enhancement network.
In the second stage, we refine the learnable prompts by utilizing backlit images, enhanced results, and well-lit images through rank learning.
The refined prompts can be used to fine-tune the enhancement network for further performance improvement.
We alternate the prompt refinement and fine-tuning of the enhancement network until we achieve visually pleasing results.
It should be noted that the CLIP model remains fixed throughout the learning process, and our method does not introduce any additional computational burden apart from prompt initialization and refinement.
We provide further details on the key components of our approach below.

\subsection{Initial Prompts and Enhancement Training}
\label{subsec:initial_prompt}

The first stage of our approach involves the initialization of negative and positive (learnable) prompts to roughly characterize backlit and well-lit images, as well as the training of the initial enhancement network.

\noindent
\textbf{Prompt Initialization.}
The process of prompt initialization is depicted in Fig.~\ref{fig:iterative_prompt_learning_framework}(a). 
Given a backlit image $I_{b} \in \mathbb{R}^{H\times W \times 3}$ and a well-lit image $I_{w} \in \mathbb{R}^{H\times W \times 3}$ (as reference), we randomly initialize a positive prompt $T_{p} \in \mathbb{R}^{N\times 512}$ and a negative prompt $T_{n} \in \mathbb{R}^{N\times 512}$. $N$ represents the number of embedded tokens in each prompt.
Then, we feed the backlit and well-lit images to the image encoder $\Phi_{image}$ of the pre-trained CLIP to obtain their latent code.
Meanwhile, we also extract the latent code of the positive and negative prompts by feeding them to the text encoder $\Phi_{text}$.
Based on the text-image similarity in the CLIP latent space, we use the binary cross entropy loss of classifying the backlit and well-lit images to learn the initial prompt pair: 
\begin{equation}
    \mathcal{L}_{initial}= -(y*\log(\hat{y})+(1-y)*\log(1-\hat{y})),
    \label{eq:cross_entropy}
\end{equation}
\begin{equation}
    \hat{y}=\frac{e^{cos(\Phi_{image}(I),\Phi_{text}(T_p))}}{\sum_{i\in{\{n,p\}}}e^{cos(\Phi_{image}(I),\Phi_{text}(T_i))}},
\end{equation}
%
where $I \in \{I_{b},I_{w}\}$ and $y$ is the label of the current image, $0$ is for negative sample $I_{b}$ and $1$ is for positive sample $I_{w}$.

\noindent
\textbf{Training the Initial Enhancement Network.}
Given the initial prompts obtained from the first stage, we can train an enhancement network with a CLIP-aware loss.
As a baseline model, we use a simple Unet \cite{Unet} to enhance the backlit images, though more complex networks can also be employed.
Inspired by the Retinex model \cite{Retinex}, which is widely used for light enhancement, the enhancement network estimates the illumination map $I_{i} \in \mathbb{R}^{H\times W \times 1}$ and then produces the final result via $I_{t}= I_{b}/I_{i}$.
To train the enhancement network, we employ  CLIP-Enhance loss $\mathcal{L}_{clip}$ and identity loss $\mathcal{L}_{identity}$.

The CLIP-Enhance loss measures the similarity between the enhanced result and the prompts in the CLIP space:
\begin{equation}
    \mathcal{L}_{clip}=\frac{e^{cos(\Phi_{image}(I_t),\Phi_{text}(T_n))}}{\sum_{i\in{\{n,p\}}}e^{cos(\Phi_{image}(I_t),\Phi_{text}(T_i))}}.
    \label{L-CLIP-enhance}
\end{equation}

The identity loss encourages the enhanced result to be similar to the backlit image in terms of content and structure: 
\begin{equation}
\label{eq:identity_loss}
\mathcal{L}_{identity}=\sum_{l=0}^{4}{\alpha_l}\cdot||\Phi_{image}^{{l}}(I_{b})-\Phi_{image}^{l}(I_{t})||_2,
\end{equation}
where $\alpha_l$ is the weight of the $l^{th}$ layer of the image encoder in the ResNet101 CLIP model. The final loss for training the enhancement network is the combination of the two losses:

\begin{equation}
    \mathcal{L}_{enhance}=\mathcal{L}_{clip}+w\cdot \mathcal{L}_{identity},
    \label{L-enhance}
\end{equation}
where $w$ is the weight to balance the magnitude of different loss terms and is set to $0.9$ empirically.
We divide the training schedule into two parts. First, we use the identity loss to implement self-reconstruction as it encourages the enhanced result to be similar to the backlit image in the pixel space.
Then, we use both the identity loss and the CLIP-Enhance loss to train the network.
For the identity loss, we set $\alpha_{l=0,1,...,4}$ in Eq.~\eqref{eq:identity_loss} to $1.0$ during the self-reconstruction stage. During training of the backlit enhancement network, we set $\alpha_{l=0,1,2,3}=1.0$ and $\alpha_{l=4}=0.5$. This is because we found that the features of the last layer are more related to the color of the images, which is what we want to adjust.

\subsection{Prompt Refinement and Enhancement Tuning}
\label{sebsec:second_stage}

In the second stage, we iteratively perform prompt refinement and enhancement network tuning. The prompt refinement and the tuning of the enhancement network are conducted in an alternating manner. The goal is to improve the accuracy of learned prompts for distinguishing backlit images, enhanced results, and well-lit images, as well as perceiving heterogeneous regions with different luminance.

\noindent
\textbf{Prompt Refinement.}
We observed that in some cases, using only the initial prompts obtained from the backlit and well-lit images is insufficient for enhancing the color and illuminance.
This is because the initial prompts may fail to capture the fine-grained differences among the backlit images, enhanced results, and well-lit images.
To address this, we propose a further refinement of the learnable positive and negative prompts.
Given the result $I_{t} \in \mathbb{R}^{H\times W \times 3}$ enhanced by the current enhancement network, we use a margin ranking loss to update the prompts. The process of prompt refinement is illustrated in Fig. \ref{fig:iterative_prompt_learning_framework}(b).

Formally, we define the negative similarity score between the prompt pair and an image as:
\begin{equation}
    S(I)=\frac{e^{cos(\Phi_{image}(I),\Phi_{text}(T_n))}}{\sum_{i\in{\{n,p\}}}e^{cos(\Phi_{image}(I),\Phi_{text}(T_i))}},
    \label{eq:similarity_score}
\end{equation}
Then, the margin ranking loss can be expressed as:
\begin{equation}
    \begin{aligned}
    \mathcal{L}_{prompt1}&=\max(0,S(I_{w})-S(I_{b})+m_0)\\
    &+\max(0,S(I_{t})-S(I_{b})+m_0)\\
    &+\max(0,S(I_{w})-S(I_{t})+m_1),
    \label{eq:pt3}
    \end{aligned}
\end{equation}
where $m_0 \in [0,1]$ represents the margin between the score of well-lit/enhanced results and the backlit images in the CLIP embedding space. 
We set $m_0$ to 0.9 to extend the distance between backlit images and well-lit images as much as possible. 
Meanwhile, $m_1$ represents the margin between the score of the enhanced results and the well-lit images in the CLIP embedding space. We set $m_1$ to 0.2 to ensure that the enhanced results are similar to well-lit images. These hyperparameters are chosen empirically and there will be more explanation in supplementary material. 
%

To ensure that the iterative learning can improve the performance in each iterative round, we preserve the previous enhanced results $I_{t-1}$ obtained by the previous enhancement network in the ranking process.
We add the two groups of enhanced results, $I_{t-1}$ and $I_{t}$, into the constraints, enabling the newly learned prompts to focus more on the light and color distribution of images, rather than high-level content in the image (see Fig.~\ref{fig:sequence of photos}).
The loss function is modified as:
\begin{equation}
    \begin{aligned}
    \mathcal{L}_{prompt2}&=\max(0,S(I_{w})-S(I_{b})+m_0)\\&+\max(0,S(I_{t-1})-S(I_{b})+m_0)\\&+\max(0,S(I_{w})-S(I_{t})+m_1)\\&+\max(0,S(I_{t})-S(I_{t-1})+m_2),
    \end{aligned}
    \label{eq:pt4}
\end{equation}

\noindent where $m_2$ represents the margin between the newly enhanced results and previously enhanced results. We set $m_2=m_1$ as the margins $m_1$ and $m_2$ have the same target, keeping the two image groups similar.

\begin{figure}[!t]
    \begin{center}
           \includegraphics[width=1.0\linewidth]{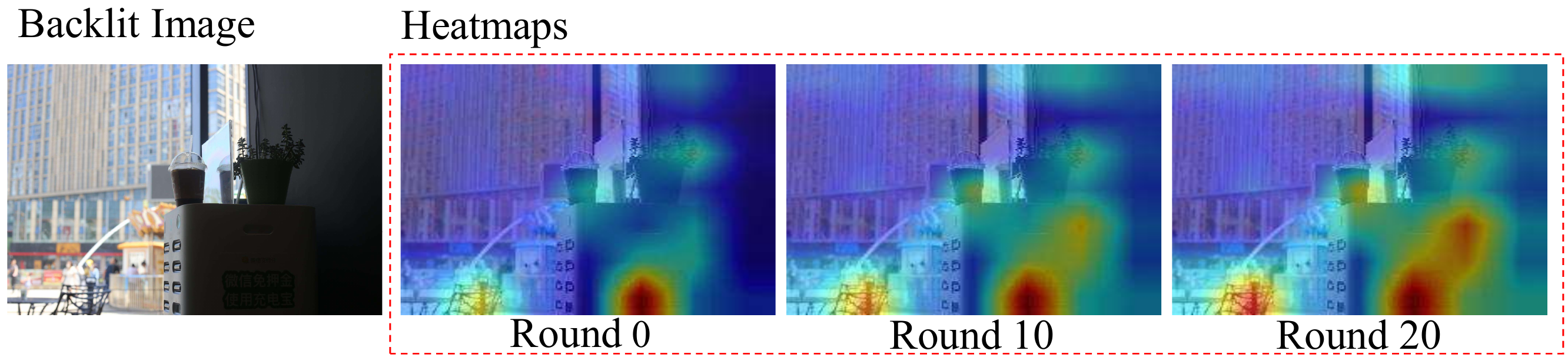}
    \end{center}
    \vspace{-1.8em}
    \caption{Attention map changes with iterative learning.}
    \label{fig:heatmap}
    \vspace{-1em}
\end{figure}

\begin{figure}[!t]
    \begin{center}
           \includegraphics[width=1.0\linewidth]{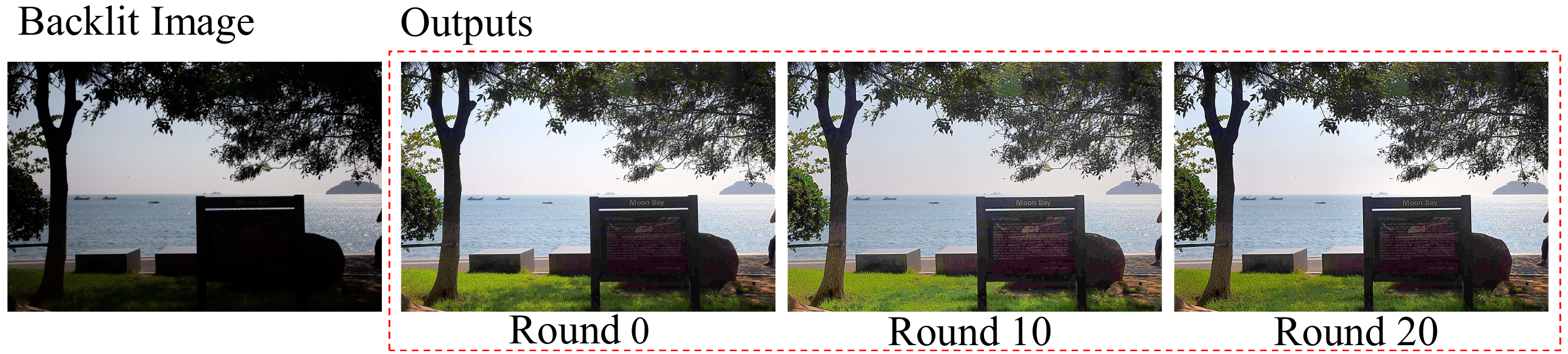}
    \end{center}
    \vspace{-1.8em}
    \caption{Enhanced results of different iteration rounds.}
    \label{fig:visual_iter}
    \vspace{-2em}
\end{figure}

\noindent
\textbf{Tuning the Enhancement Network.} The tuning of the enhancement network follows the same process in Sec.~\ref{subsec:initial_prompt} except we use the refined prompts to compute for the CLIP-Enhance loss $\mathcal{L}_{clip}$ and  generate the enhanced training data from the updated network to further refine the prompt.

\begin{figure*}[!t]
\vspace{-1em}
 \begin{center}
  \begin{tabular}{c@{ }c@{ }c@{ }c@{ }c@{ }c@{ }c@{ }}
   \includegraphics[height=5cm]{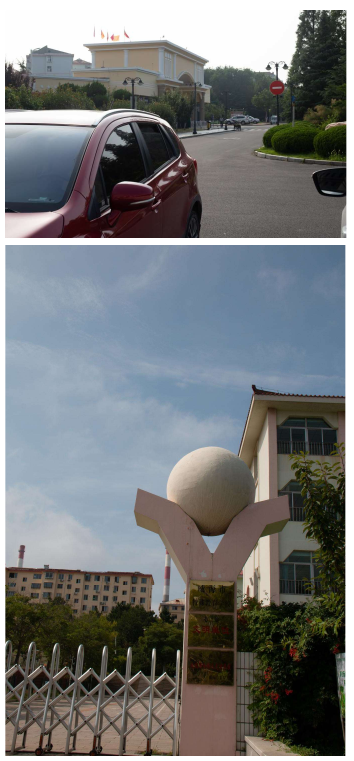}&~
   \hspace{-0.7em}
   \includegraphics[height=5cm]{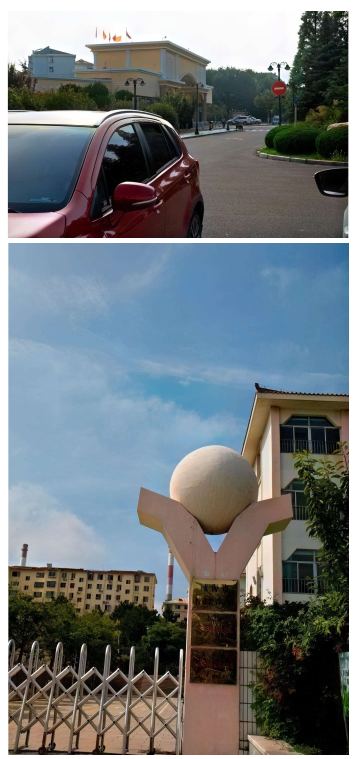}&~
   \hspace{-0.7em}
   \includegraphics[height=5cm]{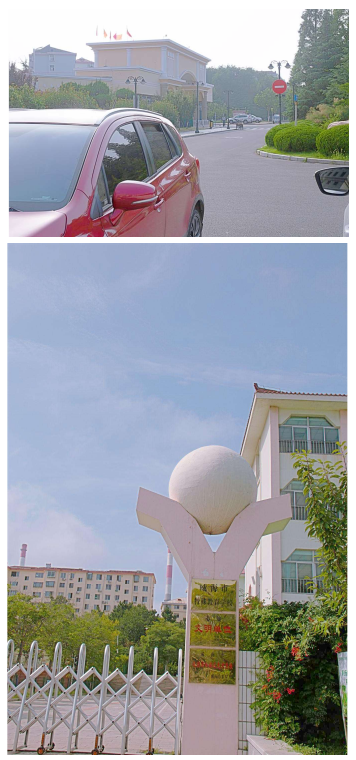}&~
   \hspace{-1.7em}
   \includegraphics[height=5cm]{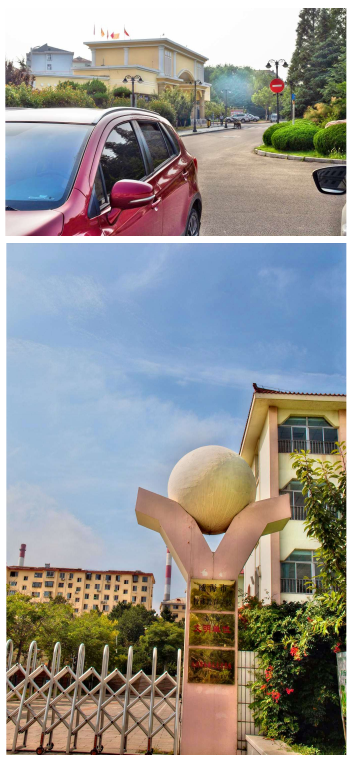}&~
   \hspace{-1.7em}
   \includegraphics[height=5cm]{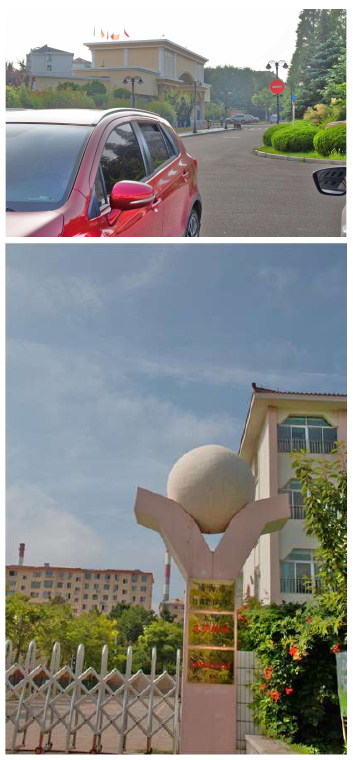}&~
   \hspace{-0.7em}
   \includegraphics[height=5cm]{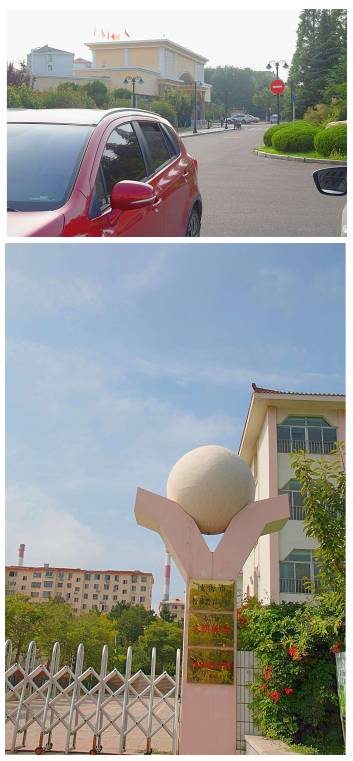}&~
   \hspace{-0.7em}
   \includegraphics[height=5cm]{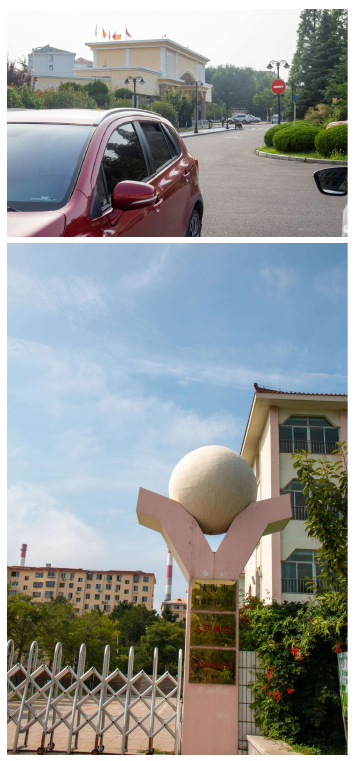}\\
    \small{Input}  &~ \small{Afifi et al.}~\cite{Mahoud2021}&~ \small{Zero-DCE}~\cite{Guo2020CVPR}&~ \small{EnlightenGAN}~\cite{Jiang2019}&~ \small{ExCNet}~\cite{zhang2019zero}&~  \small{CLIP-LIT (Ours)} &~ \small{Ground Truth}\\
  \end{tabular}
 \end{center}
 \vspace{-1.8em}
 \caption{Visual comparison on the backlit images sampled from the BAID test dataset. }
 \label{fig:comparison}
 \vspace{-1.8em}
\end{figure*}

\noindent
\textbf{Discussion.} 
To show the effectiveness of iterative learning, following Chefer~\etal~\cite{chefer2021generic}, we visualize the attention maps in the CLIP model for the interaction between the learned negative prompt and an input image at different alternate rounds. The heatmap, as shown in Fig. \ref{fig:heatmap}, represents the relevance between each pixel in the image and the learned prompt.
The heatmap shows that during iterations, the learned negative prompt becomes increasingly relevant to the regions with unpleasant lighting and color. 
We also show the enhanced results with different iterative rounds in Fig. \ref{fig:visual_iter}. At the intermediate round, the color in some enhanced regions of the outputs is over-saturated. After enough iterations, the over-saturation is corrected while the dark regions are closer to the well-lit state compared with the previous outputs. The observation here suggests the capability of our approach in perceiving heterogeneous regions with different luminance.
We will provide the quantitative comparison in Sec.~\ref{sebsec:ablation}.

\vspace{-0.4em}
\section{Experiments}
\label{Experiment}

\noindent\textbf{Dataset.}
For training, we randomly select 380  backlit images from BAID~\cite{LV2022103403} training dataset as input images and select 384 well-lit images from DIV2K~\cite{Agustsson_2017_CVPR_Workshops} dataset as reference images. 
%
%
We test our methods on the BAID test dataset, which includes 368 backlit images taken in diverse light scenarios and scenes.
To examine the generalization ability, we collected a new evaluation dataset, named Backlit300, which consists of 305 backlit images from Internet, Pexels, and Flickr. The data will be made available. 

\noindent\textbf{Training.}
We implement our method with PyTorch on a single NVIDIA GTX 3090Ti GPU. We use Adam optimizer with $\beta_1=0.9$ and $\beta_2=0.99$. 
The number $N$ of embedded tokens in each learnable prompt is set to $16$.
We set the total training iterations to $50K$, within which, the number of self-reconstruction iterations is set to $1K$, the number of prompt pair initialization learning iterations is set to $10K$. 
We set the learning rate for the prompt initialization/refinement and enhancement network training to $5\cdot 10^{-6}$ and $2\cdot10^{-5}$.
The batch size  for prompt initialization/refinement and enhancement network training is set to $8$ and $16$. 
During training, we resize the input images to $512\times512$ and use flip, zoom, and rotate as augmentations.

\noindent\textbf{Inference.}
The sizes of some input images from the BAID and Backlit300 test datasets are large, and some methods are unable to handle such high-resolution images directly.
To ensure a fair comparison, we resize all test images to have a long side of 2048 pixels if their size is larger than $2048\times2048$.

\noindent\textbf{Compared Methods.}
As there are very few publicly available deep learning-based methods for backlit image enhancement, we compare our approach with representative methods that solve related tasks, including low-light image enhancement methods such as Zero-DCE~\cite{Guo2020CVPR}, Zero-DCE++~\cite{ZeroDCE++}, SCI~\cite{SCI2022}, URetinex-Net~\cite{URetinexNet}, SNR-Aware~\cite{SNR2022}, Zhao et al.~\cite{INN}, and EnlightenGAN~\cite{Jiang2019}; exposure correction methods such as Afifi et al.~\cite{Mahoud2021}; and backlit enhancement methods such as ExCNet~\cite{zhang2019zero}.
Some methods provide different models trained on different datasets. We compare our method with all released models of different methods to ensure a fair comparison.
To further validation, we also provide retrained supervised methods' results in supplementary material.
For unsupervised methods, we retrained them on the same training data as our method to ensure that they are evaluated under the same conditions.

\begin{figure*}[!t]
\setlength\tabcolsep{1pt}
 \begin{center}
 \vspace{-1em}
 
  \begin{tabular}{c@{ }c@{ }c@{ }c@{ }c@{ }c@{ }}
  \includegraphics[height=3.6cm]{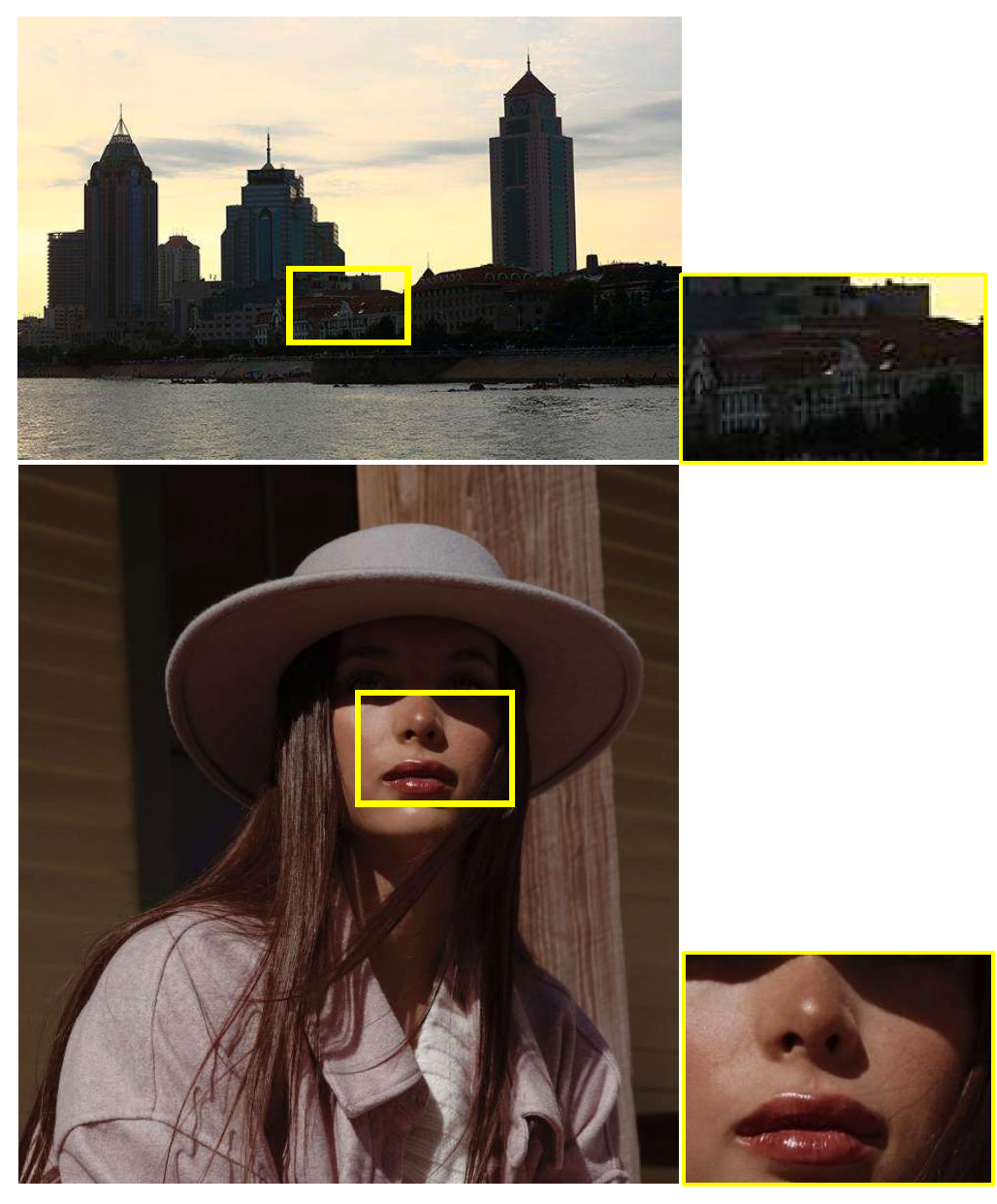}&
  \hspace{-0.5em}
   \includegraphics[height=3.6cm]{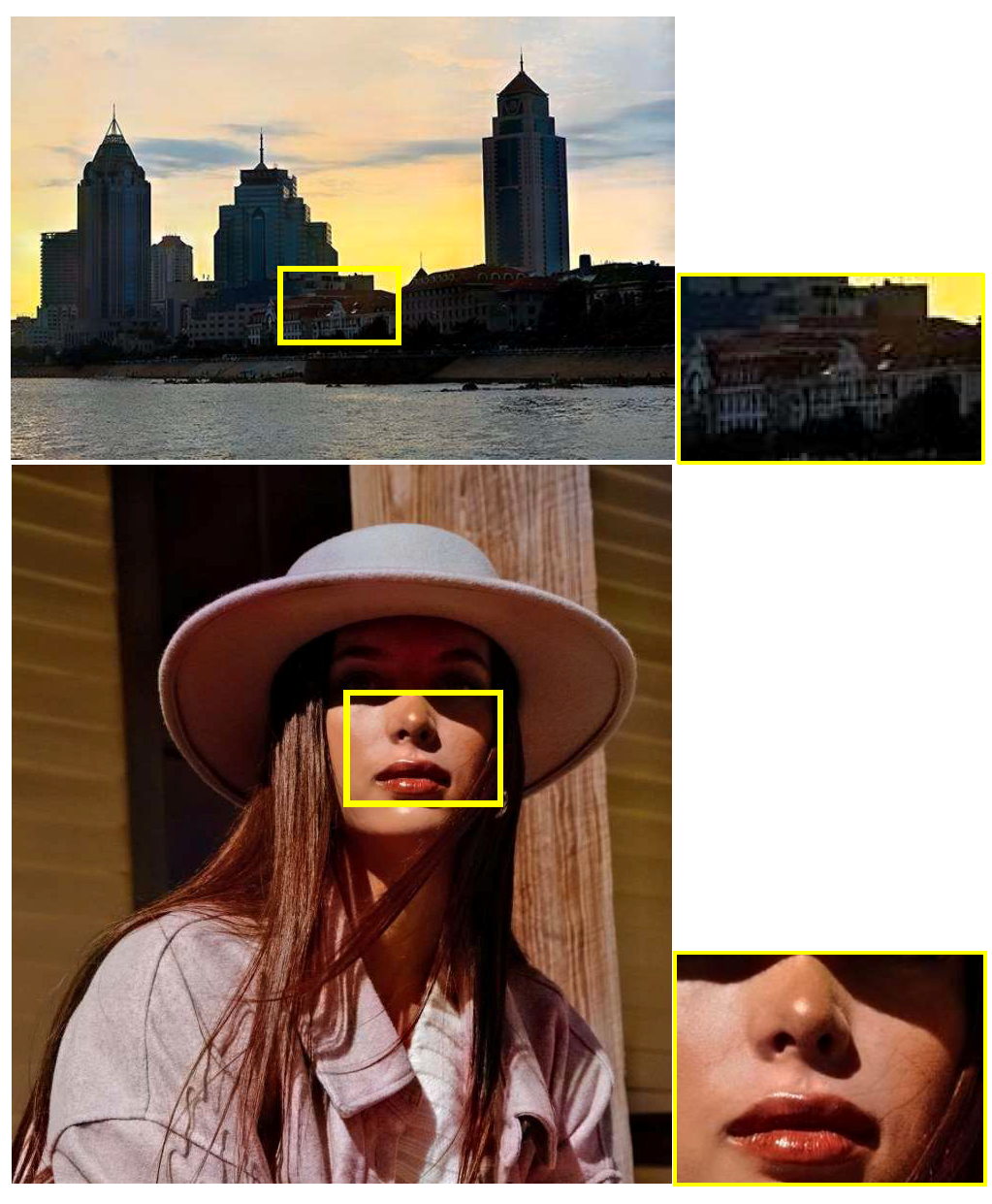}&
   \hspace{-0.5em}
   \includegraphics[height=3.6cm]{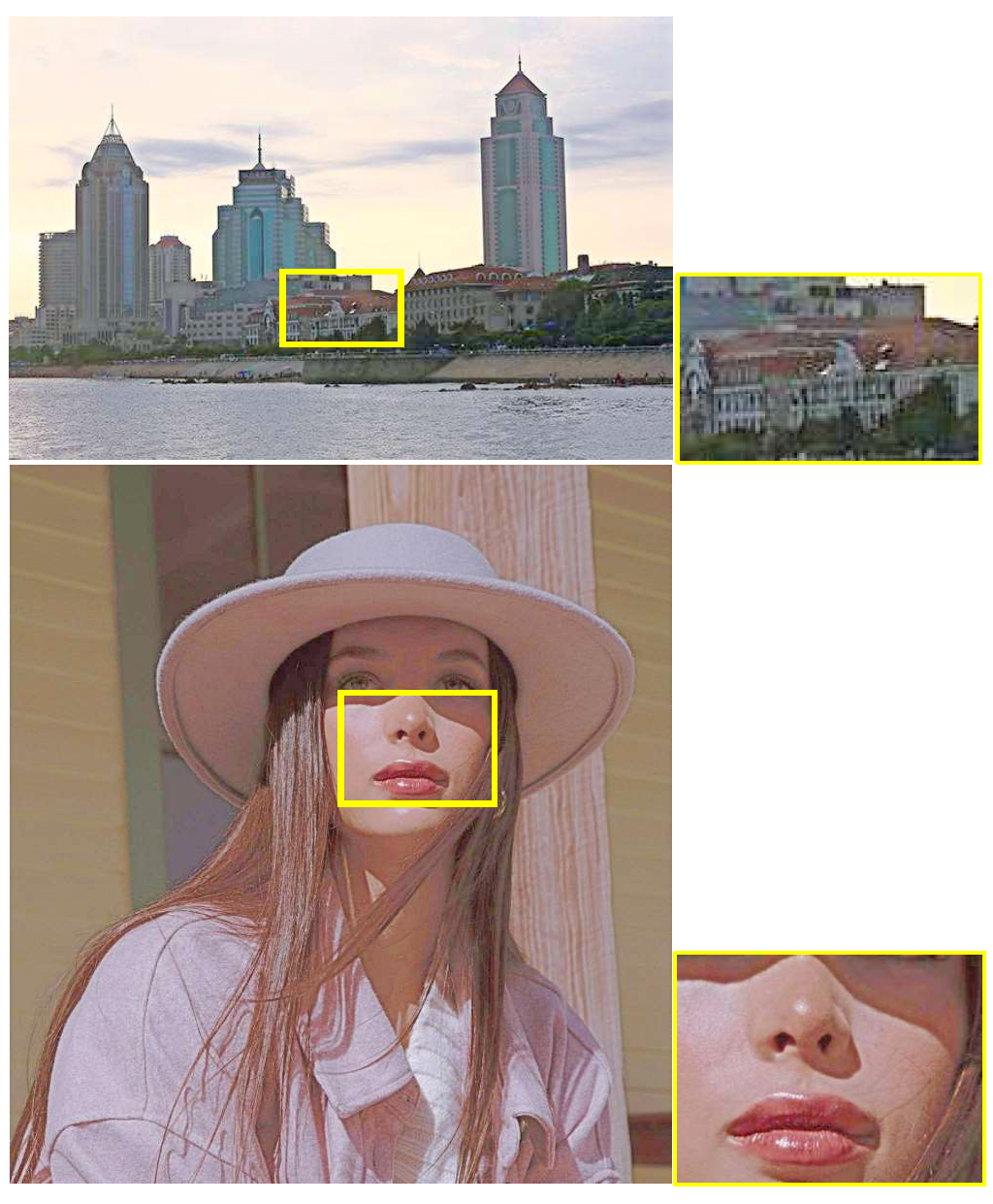}&
   \hspace{-0.5em}
   \includegraphics[height=3.6cm]{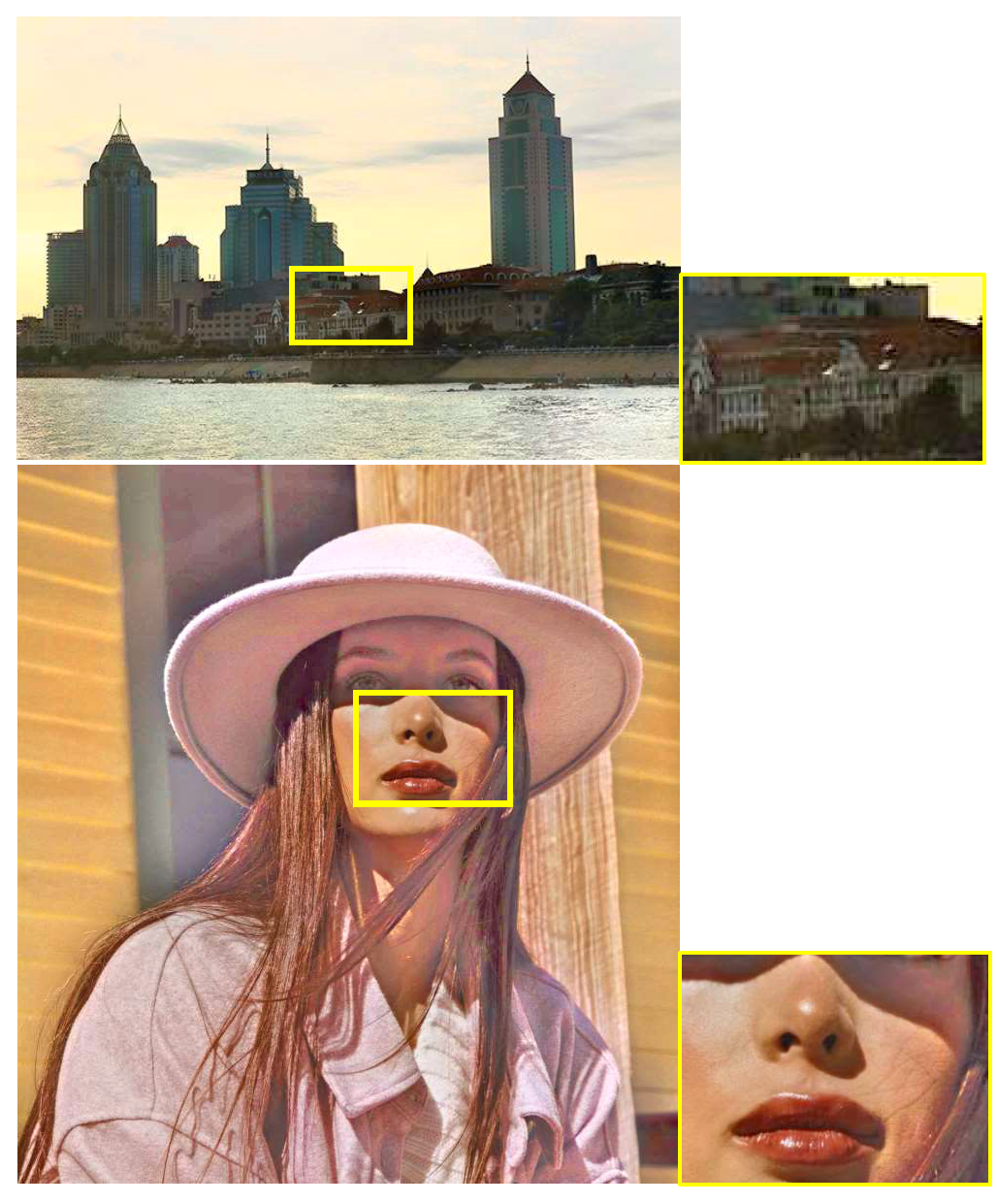}&
   \hspace{-0.5em}
   \includegraphics[height=3.6cm]{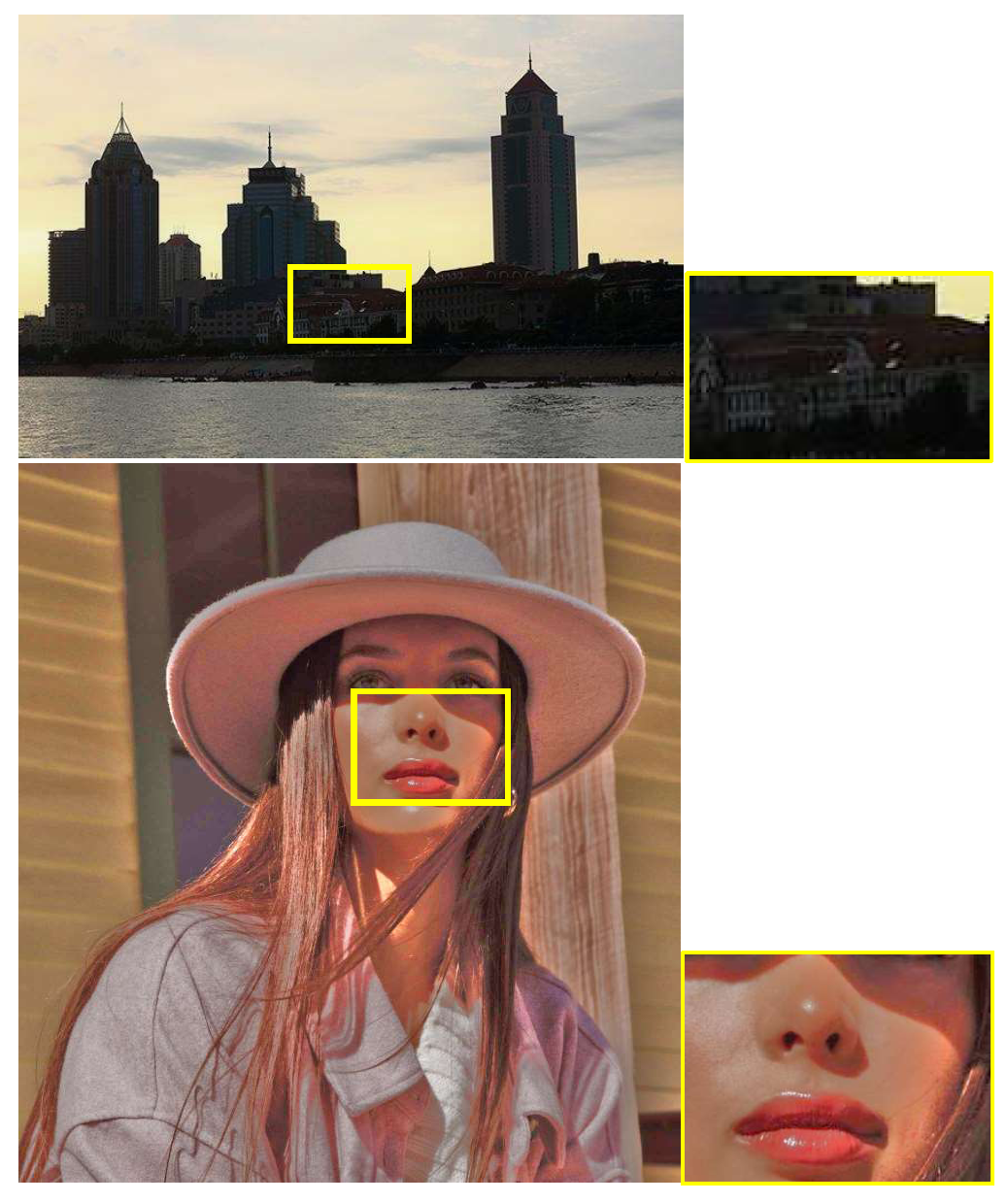}&
   \hspace{-0.5em}
   \includegraphics[height=3.62cm]{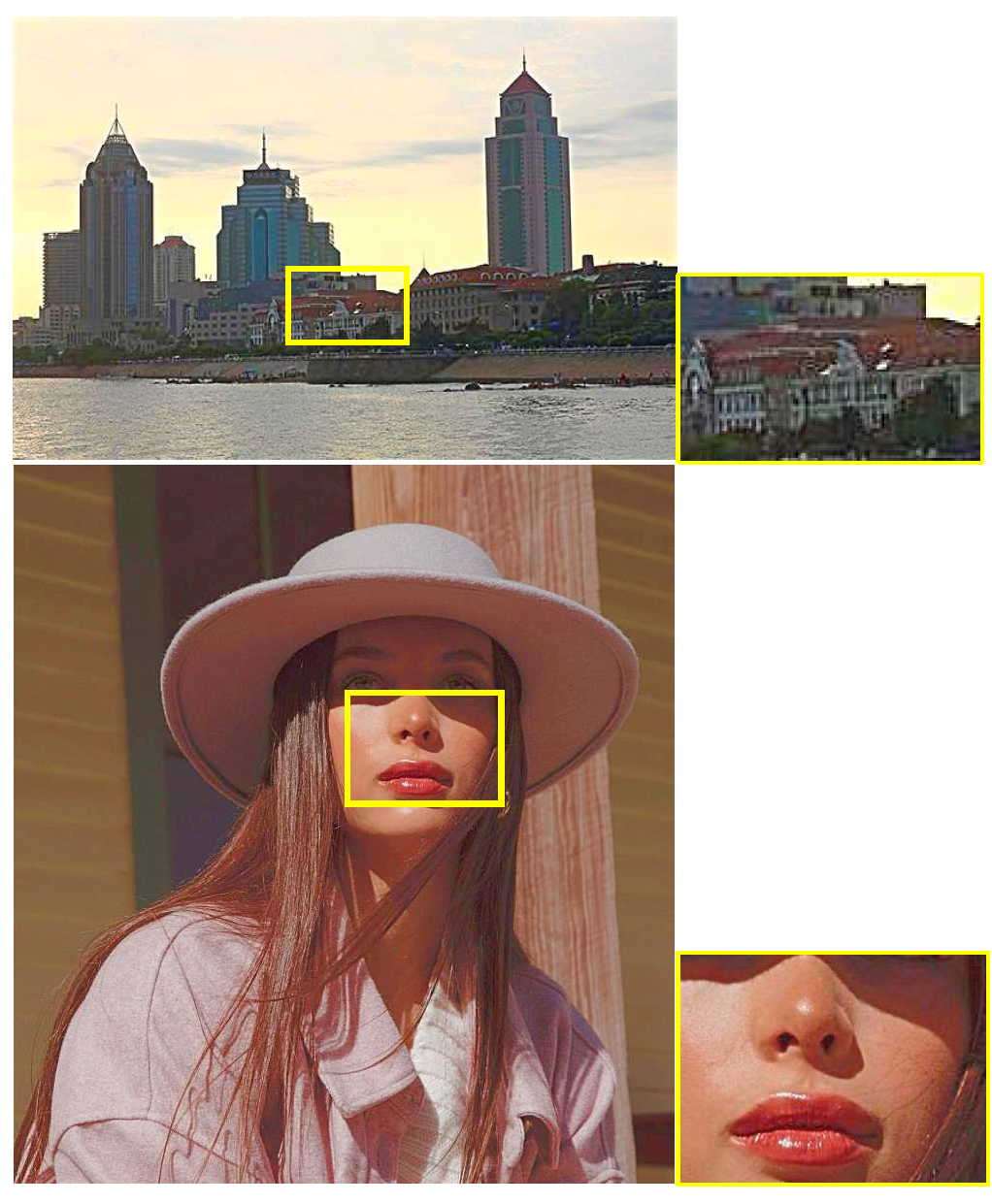}\\
    \small{Input}  & \small{Afifi et al.}~\cite{Mahoud2021}& \small{Zero-DCE}~\cite{Guo2020CVPR}& \small{EnlightenGAN}~\cite{Jiang2019}& \small{ExCNet}~\cite{zhang2019zero}&  \small{CLIP-LIT (Ours)} \\ 
  \end{tabular}
 \end{center}
 \vspace{-2em}
 \caption{Visual comparison on the backlit images sampled from the Backlit300 test dataset.}
 \label{fig:extra data comparison}
\end{figure*}

\newcommand{\best}[1]{\textcolor{red}{\textbf{#1}}}
\newcommand{\second}[1]{\textcolor{blue}{\textbf{#1}}}

\begin{table*}[!h]
\begin{minipage}{0.663\linewidth}
\vspace{-1em}
\centering
	\caption{Quantitative comparison on the BAID test dataset.  The best and second performance are marked in \textcolor{red}{red} and \textcolor{blue}{blue}.}
	\vspace{-2em}
	\begin{center}
		\resizebox{1.0\linewidth}{!}{
		\begin{tabular}{c|c|c|c|c||c}
			\hline
 			\multirow{2}{*}{Type}
			&Methods & PSNR$\uparrow$ & SSIM$\uparrow$ & LPIPS$\downarrow$  & MUSIQ$\uparrow$   \\ 
 			\cline{2-6}
			&Input &16.641   &0.768   &0.197    &   52.115      \\
 			\hline
 			\multirow{7}{*}{Supervised}
 			&Afifi et al.  \cite{Mahoud2021} &15.904 &0.745 &0.227 &52.863  \\ 
 			&Zhao et al.-MIT5K  \cite{INN} &18.228 &0.774 &0.189 &51.457  \\ 
			&Zhao et al.-LOL  \cite{INN} &17.947 &0.822 &0.272 &49.334  \\ 
			&URetinex-Net  \cite{URetinexNet} &18.925 &0.865 &0.211 &\second{54.402}  \\ 
			&SNR-Aware-LOLv1  \cite{SNR2022} &15.472 &0.747 &0.408 &26.425  \\ 
			&SNR-Aware-LOLv2real  \cite{SNR2022} &17.307 &0.754 &0.398 &26.438 \\ 
			&SNR-Aware-LOLv2synthetic  \cite{SNR2022} &17.364 &0.752 &0.403 &23.960  \\ 
			\hline
			\multirow{10}{*}{Unsupervised}
			&Zero-DCE  \cite{Guo2020CVPR} &\second{19.740} &0.871 &0.183 &51.804  \\ 
			&Zero-DCE++  \cite{ZeroDCE++} &19.658 &\best{0.883} &0.182 &48.573  \\ 
			&RUAS-LOL  \cite{RUAS2021} &9.920 &0.656 &0.523 &37.207  \\ 
			&RUAS-MIT5K  \cite{RUAS2021} &13.312 &0.758 &0.347 &45.008  \\ 
			&RUAS-DarkFace  \cite{RUAS2021} &9.696 &0.642 &0.517 &39.655  \\ 
			&SCI-easy  \cite{SCI2022} &17.819 &0.840 &0.210 &51.984  \\ 
			&SCI-medium  \cite{SCI2022} &12.766 &0.762 &0.347 &44.176 \\ 
			&SCI-diffucult  \cite{SCI2022} &16.993 &0.837 &0.232 &52.369 \\ 

			&EnlightenGAN  \cite{Jiang2019} &17.550 &0.864 &0.196 &48.417 \\ 
			
			&ExCNet \cite{zhang2019zero} &19.437 &0.865 &\second{0.168} &52.576  \\ 
			\hline
 			\multirow{5}{*}{\shortstack{Unsupervised\\(retrained)}}
			&Zero-DCE  \cite{Guo2020CVPR} &18.553 &0.863 &0.194 &49.436  \\ 
			&Zero-DCE++  \cite{ZeroDCE++} &16.018 &0.832 &0.240 &47.253  \\ 
			&RUAS  \cite{RUAS2021} &12.922 &0.743 &0.362 &45.056  \\ 
			&SCI \cite{SCI2022} &16.639 &0.768 &0.197 &52.265 \\
			&EnlightenGAN  \cite{Jiang2019} &17.957 &0.849 &0.182 &53.871 \\ 
			&CLIP-LIT (Ours) &\best{21.579} &\best{0.883} &\best{0.159} &\best{55.682}  \\
			\hline
		\end{tabular}
		}
	\end{center}
	\label{table:quantitative}
	\vspace{-2em}
\end{minipage}\quad
\begin{minipage}{0.337\linewidth}
\vspace{-1em}
\centering
	\caption{Quantitative comparison on the Backlit300 test dataset.}
	\vspace{-2em}
	\begin{center}
		\resizebox{1.0\linewidth}{!}{
		\begin{tabular}{c|c}
			\hline
			Methods & MUSIQ$\uparrow$   \\ 
 			\hline
			Input &51.900     \\
 			\hline
 			
 			Afifi et al.  \cite{Mahoud2021} &51.930  \\ 
 			Zhao et al.-MIT5K  \cite{INN} &50.354  \\ 
			Zhao et al.-LOL  \cite{INN} &48.334  \\ 
            URetinex-Net  \cite{URetinexNet} &51.551  \\ 
			SNR-Aware-LOLv1  \cite{SNR2022} &29.915  \\ 
			SNR-Aware-LOLv2real  \cite{SNR2022} &30.903 \\ 
			SNR-Aware-LOLv2synthetic  \cite{SNR2022} &29.149  \\ 
			\hline
			Zero-DCE  \cite{Guo2020CVPR} &51.250  \\ 
			Zero-DCE++  \cite{ZeroDCE++} &48.435 \\ 
			RUAS-LOL  \cite{RUAS2021} &40.329 \\ 
			RUAS-MIT5K  \cite{RUAS2021} &44.523  \\ 
			RUAS-DarkFace  \cite{RUAS2021} &38.934  \\ 
			SCI-easy  \cite{SCI2022} &50.642  \\ 
			SCI-medium  \cite{SCI2022} &43.493 \\ 
			SCI-diffucult  \cite{SCI2022} &49.428 \\ 
			
			EnlightenGAN  \cite{Jiang2019} &48.308 \\ 

			ExCNet \cite{zhang2019zero} &50.278  \\ 
			\hline
			Zero-DCE  \cite{Guo2020CVPR} &48.491  \\ 
			Zero-DCE++  \cite{ZeroDCE++} &46.000  \\ 
			RUAS  \cite{RUAS2021} &45.251  \\ 
			SCI \cite{SCI2022} &\second{51.960} \\
			EnlightenGAN  \cite{Jiang2019} &48.261 \\ 
			CLIP-LIT (Ours) &\best{52.921}  \\
			\hline
		\end{tabular}
		}
	\end{center}
	\label{table:extra_quantitative}
	\vspace{-2em}
\end{minipage}
\end{table*}

\vspace{-0.4em}
\subsection{Results}
\label{sec:exp_comparison}

\noindent\textbf{Visual Comparison.}
We present visual comparisons of some typical samples from the BAID test dataset in Fig. \ref{fig:comparison}. Due to space limitations, we only show the results of the best-performing methods. The complete comparisons of all methods can be found in the supplementary material. Our method consistently produces visually pleasing results with improved color and luminance without over- or under-exposure. Moreover, our method excels in handling challenging backlit regions, restoring clear texture details and satisfactory luminance without introducing any artifacts, while other methods may either fail to address such regions or produce unsatisfactory results with visible artifacts.

We also evaluate our method on the Backlit300 test dataset, and present the comparison results in Fig. \ref{fig:extra data comparison}. We can see that compared to EnlightenGAN \cite{Jiang2019} and ExCNet \cite{zhang2019zero}, our method produces results without visible distortion artifacts. Our method is also more effective in enhancing dark regions, unlike Afifi et al. \cite{Mahoud2021} and EXCNet \cite{zhang2019zero}. Moreover, our results exhibit better color contrast and input-output consistency in well-lit regions. We emphasize that our method achieves these results without the need for paired data, which is not available in many real-world scenarios.

\noindent\textbf{Quantitative Comparison.}
We use three full-reference image quality evaluation (IQA) metrics, i.e., PSNR, SSIM \cite{SSIM}, and LPIPS \cite{LPIPS}  (Alex version) and one non-reference IQA metric MUSIQ \cite{MUSIQ} to evaluate the quantitative results. 
As current non-reference IQA metrics only evaluate the overall image quality, they may not accurately measure the results of backlit image enhancement. Hence, we primarily rely on the state-of-the-art MUSIQ metric to evaluate the performance.

The quantitative comparison on the BAID test dataset is presented in Tab.~\ref{table:quantitative}. Our method outperforms all state-of-the-art methods in terms of the full-reference IQA metrics, indicating that the results generated by our method preserve the content and structure of the original images well and are close to the reference images retouched by photographers. Our method also performs the best in the non-reference MUSIQ metric when compared to other methods, demonstrating the good image quality of our results. We also report the quantitative comparison on the Backlit300 test dataset in Tab.~\ref{table:extra_quantitative}, where our method continues to achieve the best performance, further indicating the effectiveness of our method.

\begin{figure*}[!t]
     \vspace{-0.2em}
    
    \begin{center}
    
           \includegraphics[width=1.0\linewidth]{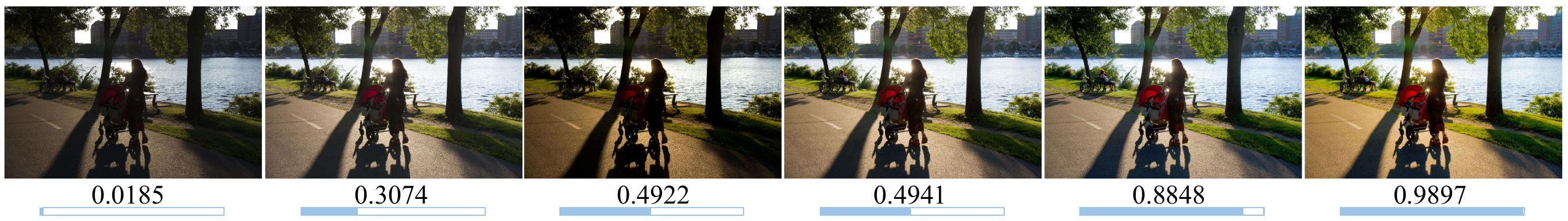}
    \end{center}
    \vspace{-1.8em}
    \caption{Comparison of similarity scores: learned positive prompt vs. the same images with gradually improved luminance and color conditions, indicating the learned prompts' sensitivity to light and color distribution rather than high-level content.}
    \label{fig:sequence of photos}
    \vspace{-1.5em}
\end{figure*}

\vspace{-0.4em}
\subsection{User Study}
\label{sec:user_study}
\vspace{-0.3em}
We conducted a user study to more comprehensively evaluate the visual quality of enhanced results obtained by different methods. In addition to our results, we chose the results obtained from the top-3 PSNR methods: Zero-DCE \cite{Guo2020CVPR}, EXCNet \cite{zhang2019zero}, and URetinex \cite{URetinexNet}, as well as EnlightenGAN \cite{Jiang2019} since it is a related work to our method. We randomly selected 20 images from the Backlit300 test partition as the evaluation set. For each image, we provided the input backlit image, the corresponding images enhanced by our method and a baseline. A total of 40 participants were invited to select their preferred image. 

\begin{figure}
    \begin{center}
           \includegraphics[width=0.8\linewidth]{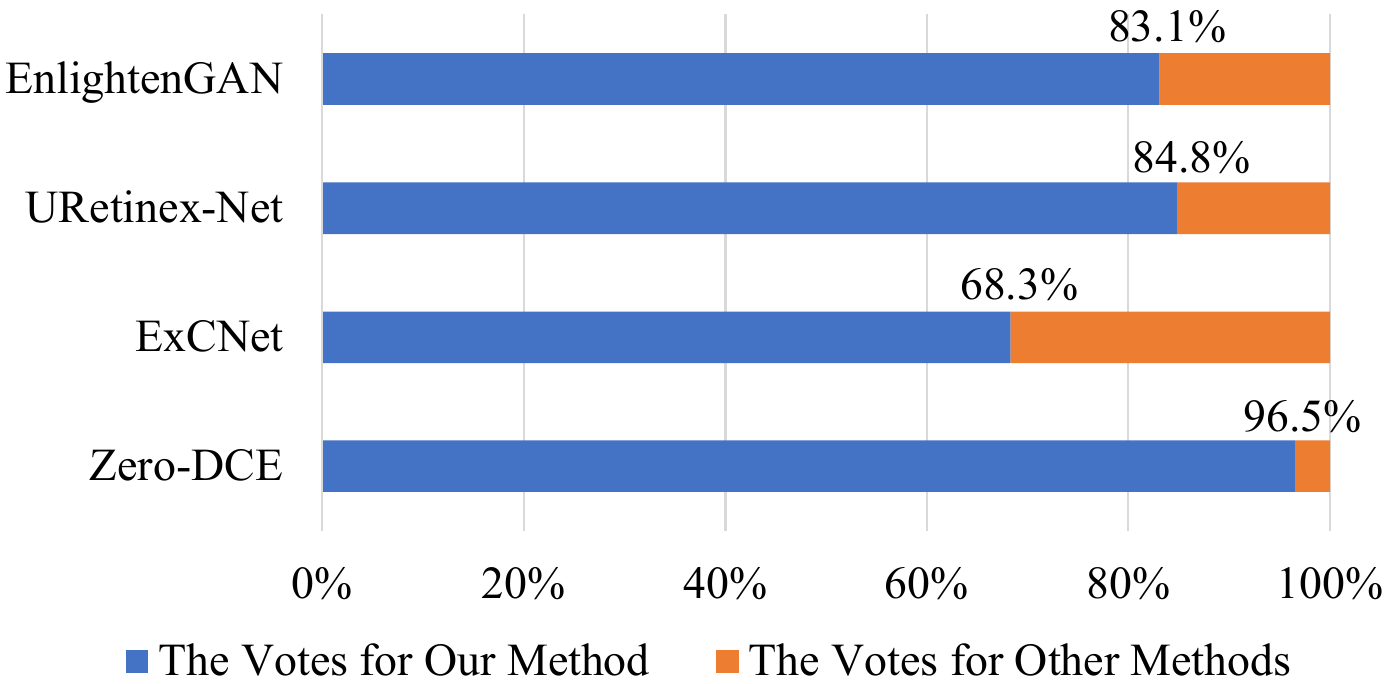}
    \end{center}
    \vspace{-1.8em}
  \caption{User Study. Voting statistics of different methods versus our method.}
  \label{fig:user_study}
  \vspace{-1em}
\end{figure}

The statistics of the user study are summarized in Fig.~\ref{fig:user_study}. The vote distribution shows that our results are the most favored by participants, with obvious advantages over the other methods. For each image, over 60\% of the participants voted for our result, indicating that our method generates more visually pleasing results when compared to other  methods.

\vspace{-0.3em}
\subsection{Ablation Studies}
\vspace{-0.3em}
\label{sebsec:ablation}

\noindent
\textbf{Effectiveness of Iterative Learning.} 
In addition to the observation provided in Sec.~\ref{sebsec:second_stage}, to further validate the effectiveness of iterative learning, we provide the quantitative comparison in Tab.~\ref{tab:three_stage_ablation}.
As presented, fine-tuning the prompts using the loss functions Eq. \eqref{eq:pt3} and  Eq. \eqref{eq:pt4} improve the enhancement performance.

\begin{table}
  \centering
    \caption{Quantitative comparisons of the iterative learning on the BAID test set.}
    \vspace{-0.8em}
  \resizebox{8cm}{!}{
  \begin{tabular}{c|c|c}
    \hline
    Method & PSNR$\uparrow$ & SSIM$\uparrow$ \\
    \hline
    fixed prompts (backlit/well-lit) &14.748 &0.823 \\ 
    w/o ranking losses (w/o Eqs. \eqref{eq:pt3} and \eqref{eq:pt4})  &20.884 &0.865 \\
    w/o $t-1$ outputs (w/o  Eq. \eqref{eq:pt4}) &20.146 &0.866 \\
    Ours &\textbf{21.579} &\textbf{0.883} \\
    \hline
  \end{tabular}
  }
  \vspace{-0.5em}
  \label{tab:three_stage_ablation}
\end{table}

\begin{figure}[t]
    \vspace{-0.5em}
    \begin{center}
        \includegraphics[width=1\linewidth]{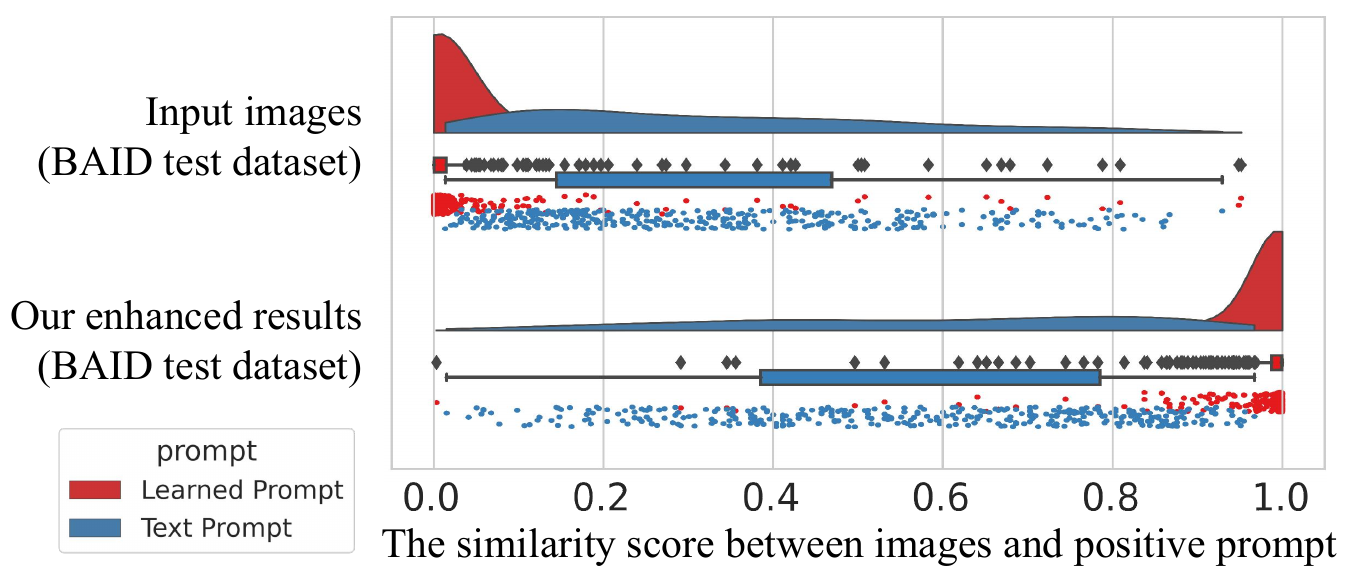}
    \end{center}
    \vspace{-1.8em}
    \caption{The distribution of similarity scores between the learned positive prompt and images across the BAID test dataset. The upper part is the kernel density estimation curve, and the lower part is the point and box plot. Our learned prompts have more precise presentation of the images' luminance quality than text prompts, such as backlit/well-lit.}
    
    \label{fig:hist}
    \vspace{-2em}
\end{figure}

\noindent
\textbf{Necessity of Prompt Refinement.} 
\label{sec:Necessity of Prompt Refinement.}
Compared to the selected words or sentences, our learned prompts can better distinguish between backlit and well-lit images (see Fig.~\ref{fig:hist}). Results in Tab.~\ref{tab:three_stage_ablation} also indicate that the enhancement model trained under the constraint of our refined prompts performs better than a fixed prompts.

\begin{figure}[!t]
    \vspace{-0.5em}
    
    \begin{center}
    \resizebox{1.0\linewidth}{!}{
           \includegraphics[width=1.0\linewidth]{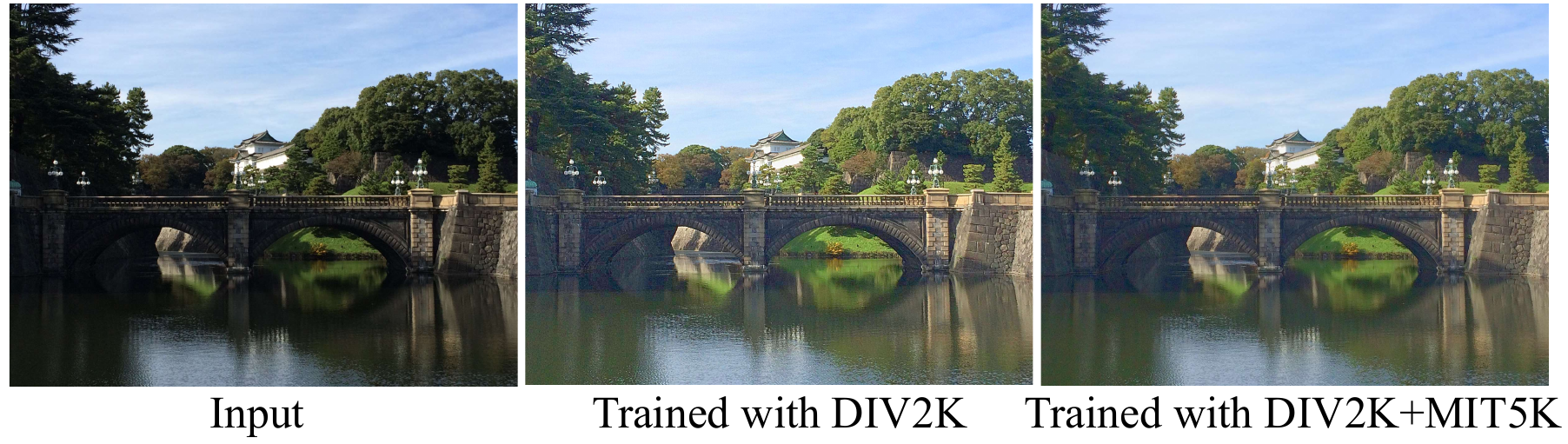}
           }
    \end{center}
    \vspace{-1.8em}
    \caption{Visual comparisons of our method trained using  different reference images.}
    \label{fig:visual_training_data_impact}
    \vspace{-1em}
\end{figure}

\noindent
\textbf{Impact of Training Data.} 
To investigate the impact of the reference data (the well-lit images) on our method, we conducted an experiment where we retrained our method on another dataset containing 1000 images selected from DIV2K\cite{Agustsson_2017_CVPR_Workshops} and MTI5K\cite{Adobe5K}, which has more diverse well-lit images. The results, as shown in Fig.~\ref{fig:visual_training_data_impact} and Tab.~\ref{tab:impact_of_training_data}, indicate that the two sets of results obtained by our method using different training data are similar, and the quantitative scores only have slight differences. Such results demonstrate that the number and variety of well-lit images used as training data have little impact on the performance of our method.

\begin{table}
  \centering
    \caption{Comparison of training data impact. The quantitative comparisons are conducted  on BAID test dataset.}
    \vspace{-0.8em}
  \resizebox{8.5cm}{!}{
  \begin{tabular}{c|c|c|c|c}
    \hline
    Reference images & PSNR$\uparrow$ & SSIM$\uparrow$ &LPIPS$\downarrow$ & MUSIQ$\uparrow$ \\
    \hline
    MIT5K~\cite{Adobe5K}+DIV2K~\cite{Agustsson_2017_CVPR_Workshops} &21.413 &\textbf{0.881} &0.162 &\textbf{56.494}\\
    DIV2K~\cite{Agustsson_2017_CVPR_Workshops}  &\textbf{21.579} &0.883 &\textbf{0.159} &55.682 \\
    \hline
  \end{tabular}
  }
  \vspace{-0.5em}
  \label{tab:impact_of_training_data}
\end{table}

\begin{table}[!t]
  \centering
  \caption{Comparison between CLIP-Enhance loss and adversarial loss on  the BAID test dataset.}
  \vspace{-1em}
  \resizebox{8.5cm}{!}{
  \begin{tabular}{c|c|c|c|c}
    \hline
    loss & PSNR$\uparrow$ & SSIM$\uparrow$ &LPIPS$\downarrow$ & MUSIQ$\uparrow$  \\
    \hline
    Adversarial loss &17.407 &0.785 &0.194 &52.416\\
    CLIP-Enhance loss &\textbf{21.579} &\textbf{0.883} &\textbf{0.159} &\textbf{55.682}  \\
    \hline
  \end{tabular}
  }
  \vspace{-1.8em}
  \label{tab:adversarial loss}
\end{table}

\noindent
\textbf{Advantage of CLIP-Enhance Loss over the Adversarial Loss.} 
To show the advantage of our CLIP-Enhance loss over the adversarial loss, we trained our enhancement network on the same unpaired training data using adversarial loss. We used the same discriminator as EnlightenGAN \cite{Jiang2019}. The results in Tab. \ref{tab:adversarial loss} indicate that our CLIP-Enhance loss achieves better enhancement performance than adversarial loss. This may be due to the fact that the CLIP prior is more sensitive to color and luminance distribution, enabling it to differentiate between images with varied lighting conditions (see Fig. \ref{fig:sequence of photos}) and perceive unbalanced luminance regions (see Fig. \ref{fig:heatmap}). 
Visual comparison is provided in supplementary material.

\vspace{-0.8em}
\section{Conclusion}
\vspace{-0.5em}

We have introduced a novel approach for training a deep network to enhance backlit images using only a few hundred unpaired data. Our method exploits the rich priors embedded in a CLIP model and leverages an iterative prompt learning strategy to generate more precise prompts that can better characterize both backlit and well-lit images. Notably, our study is the first attempt to use CLIP for low-level restoration tasks, and we anticipate that this methodology will find additional applications in the future. 

\noindent
\textbf{Acknowledgment.}
This study is supported under the RIE2020 Industry Alignment Fund Industry Collaboration Projects (IAF-ICP) Funding Initiative, as well as cash and in-kind contribution from the industry partner(s).

{\small
\bibliographystyle{ieee_fullname}
\bibliography{egbib}
}

\clearpage
\renewcommand\thesection{\Alph{section}}
\setcounter{section}{0}





\iccvfinalcopy 

\def\httilde{\mbox{\tt\raisebox{-.5ex}{\symbol{126}}}}

\ificcvfinal\pagestyle{empty}\fi




\begin{center}
      \vspace*{24pt}
      {
      \large
      \lineskip .5em
\begin{tabular}[t]{c}
	\Large\textbf{{Iterative Prompt Learning for Unsupervised Backlit Image Enhancement}}\\
\vspace{-1pt}\\
	\Large\textbf{{-- Supplementary Material --}} \\
	\vspace{15pt}
\end{tabular}
 \par
 } 
      \vskip .5em
\end{center}

\maketitle

In this supplementary material, we present more ablation studies (Section \ref{ablation}), extra training details (Section \ref{train}), further discussions (Section \ref{discussion}), more comparisons (Section \ref{comparision}, Section \ref{visual} and Section \ref{5dataset}) and discussion of extensions (Section \ref{extension}).
In addition, a video demo is provided to showcase the effectiveness of our method at \url{https://youtu.be/0qbkxNmkNWU}.

\section{More Ablation Studies}
\label{ablation}

\subsection{Superiority in Generalization Ability}
To show the superiority of our unsupervised method over the supervised method in terms of generalization ability,  we present the visual comparison with a supervised method Unet$_{pair}$  (Unet \cite{Unet} is trained using the paired training data of the BAID dataset). The visual comparisons are conducted on   the Backlit300 test dataset.

\begin{figure}[h]
\begin{center}
          \includegraphics[width=\linewidth]{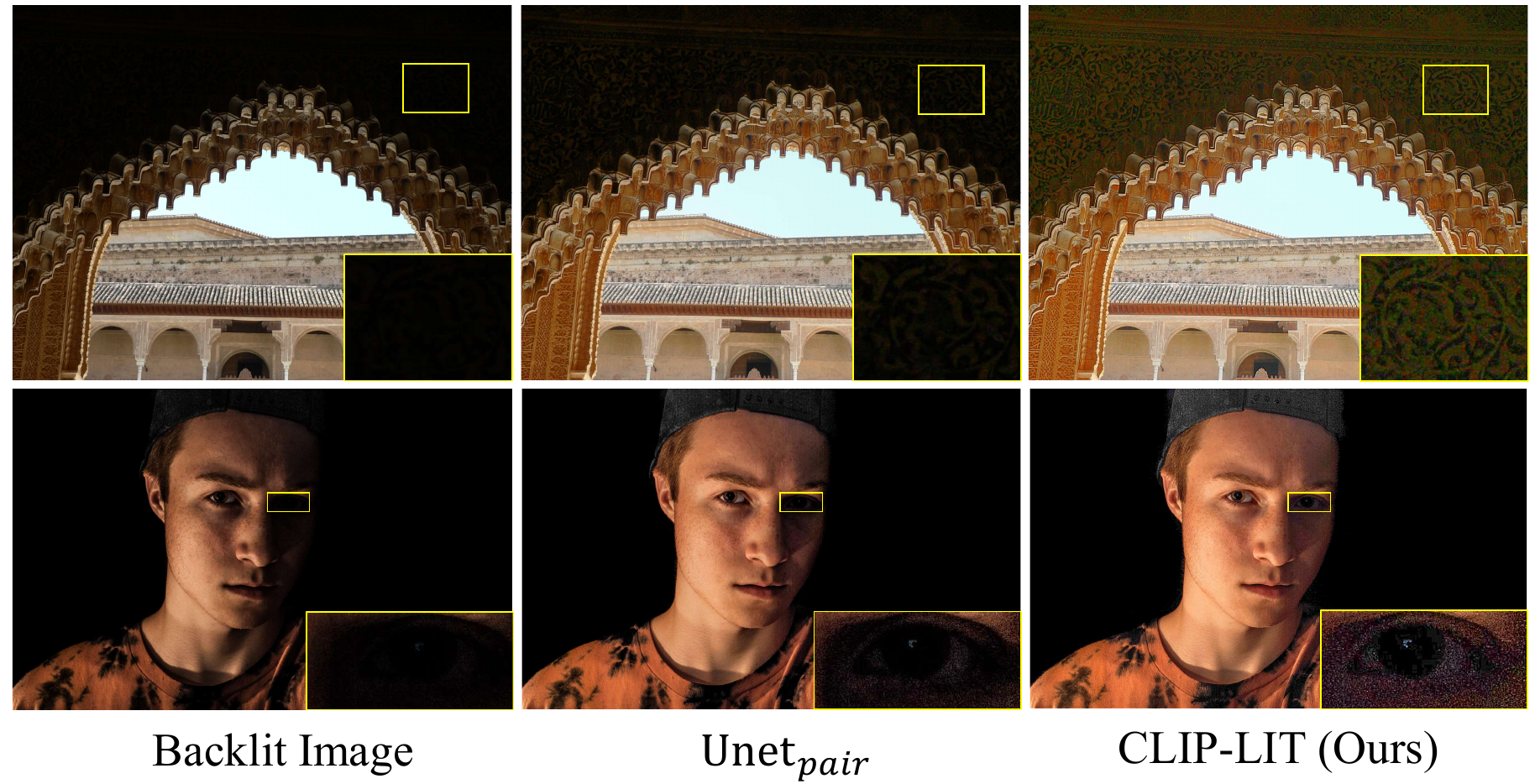}
    \end{center}
    \vspace{-1.2em}
    \caption{Visual comparisons between supervised method (Unet$_{pair}$) and our unsupervised method.}
    \vspace{-1em}
    \label{fig:visual_compare_for_genc}
\end{figure}

As shown in Fig. \ref{fig:visual_compare_for_genc}, our method produces visually pleasing results with clear details and sufficient luminance, while the results of Unet$_{pair}$ suffer from the under-exposure issue. Overall, our method yields a more realistic color distribution and a brighter image, indicating that an unsupervised backlit enhancement method is still desired even if having manually touched paired data.

\subsection{Impact of Labels on Prompt Initialization}
\begin{figure}[t]
\vspace{13.5em}
\begin{center}
          \includegraphics[width=0.7\linewidth]{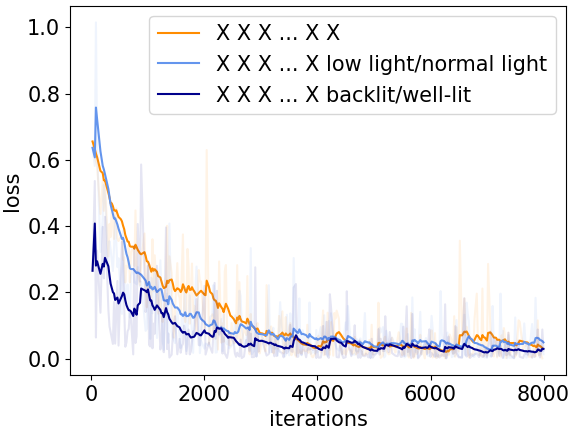}
    \end{center}
    \vspace{-1.6em}
    \caption{Prompt initialization learning investigation.}
    \label{fig:visual_prompt_init}
    \vspace{-0.5em}
\end{figure}

While Fig.~\ref{fig:visual_prompt_init} shows using pure random initialization will take more iterations to converge, both Tab.~\ref{tab:Prompt_init} and Fig.~\ref{fig:visual_prompt_init} show  whether to initialize with words does not have much impact on the final performance and model training.

\begin{table}[!t]
  \centering
  \resizebox{\linewidth}{!}{
  \begin{tabular}{c|c|c|c||c}
    \hline
    Settings& PSNR$\uparrow$ & SSIM$\uparrow$ &LPIPS$\downarrow$ &MUSIQ$\uparrow$\\
    \hline
    Pure random initialization &21.237&0.884& 0.158 & 55.959 \\
    Random initialization+backlit/well-lit  &21.527&0.882 &0.159 &55.946\\
    Random initialization+low light/normal light &21.579 &0.883 &0.159 &55.682\\
    \hline            
  \end{tabular}
  }
    \vspace{-0.6em}
  \caption{Quantitative comparison of different prompt initialization learning on BAID test dataset.}
  \label{tab:Prompt_init}
  \vspace{-0.5em}

\end{table}

\begin{figure}[!t]
\begin{center}
    \includegraphics[width=\linewidth]{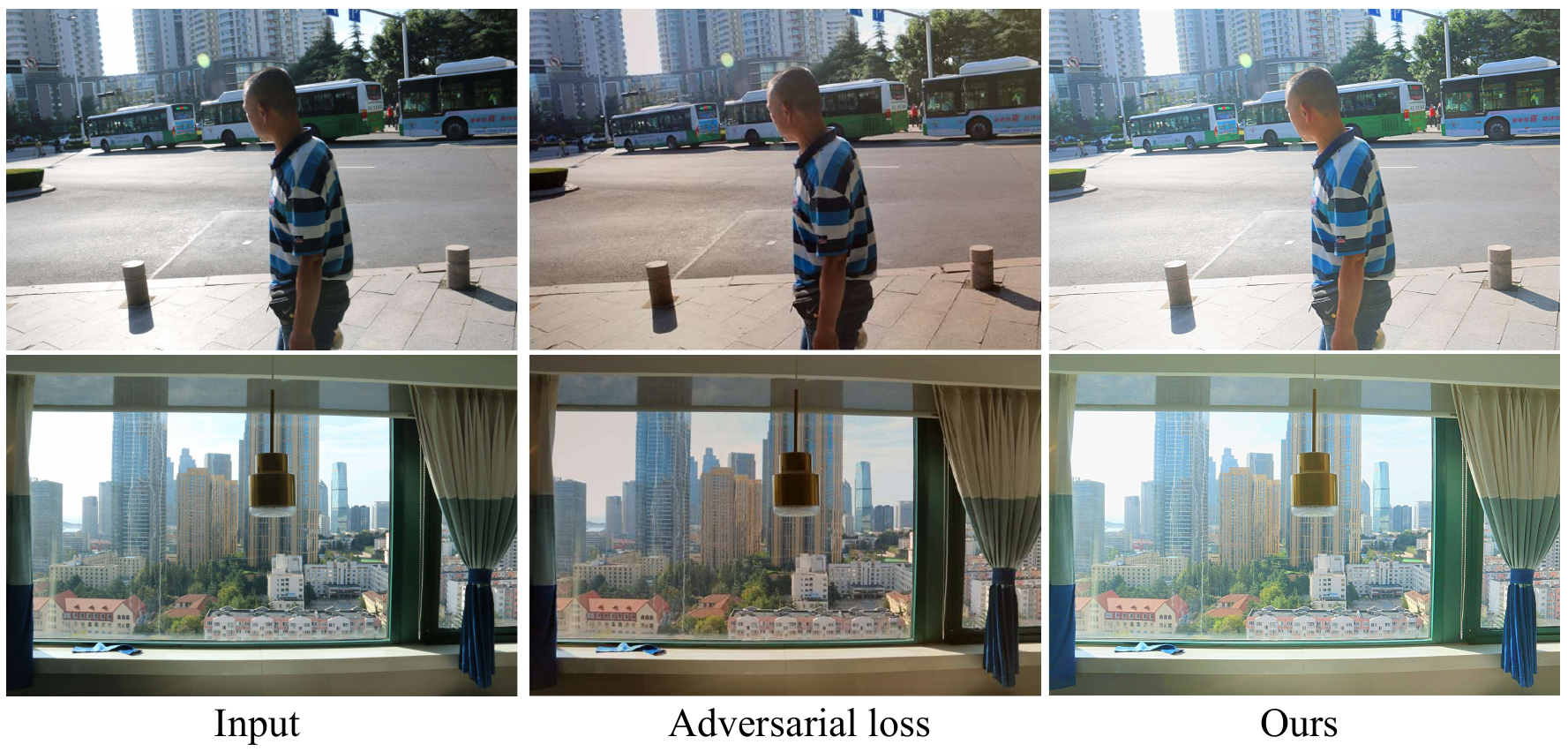}
    \end{center}
    \vspace{-1.5em}
    \caption{Visual comparisons between the model trained using adversarial loss and our method.}
    \vspace{-1.5em}
    \label{fig:visual_compare_for_adver}
\end{figure}

\subsection{Visual Comparisons with Adversarial Loss}
As shown in Fig.~\ref{fig:visual_compare_for_adver}, compared to the results generated by the model trained with adversarial loss (keeping other settings fixed), our results are more consistent with the input and brighter in backlit areas.

\begin{figure*}[t]
\vspace{1em}
\begin{center}
          \includegraphics[width=\linewidth]{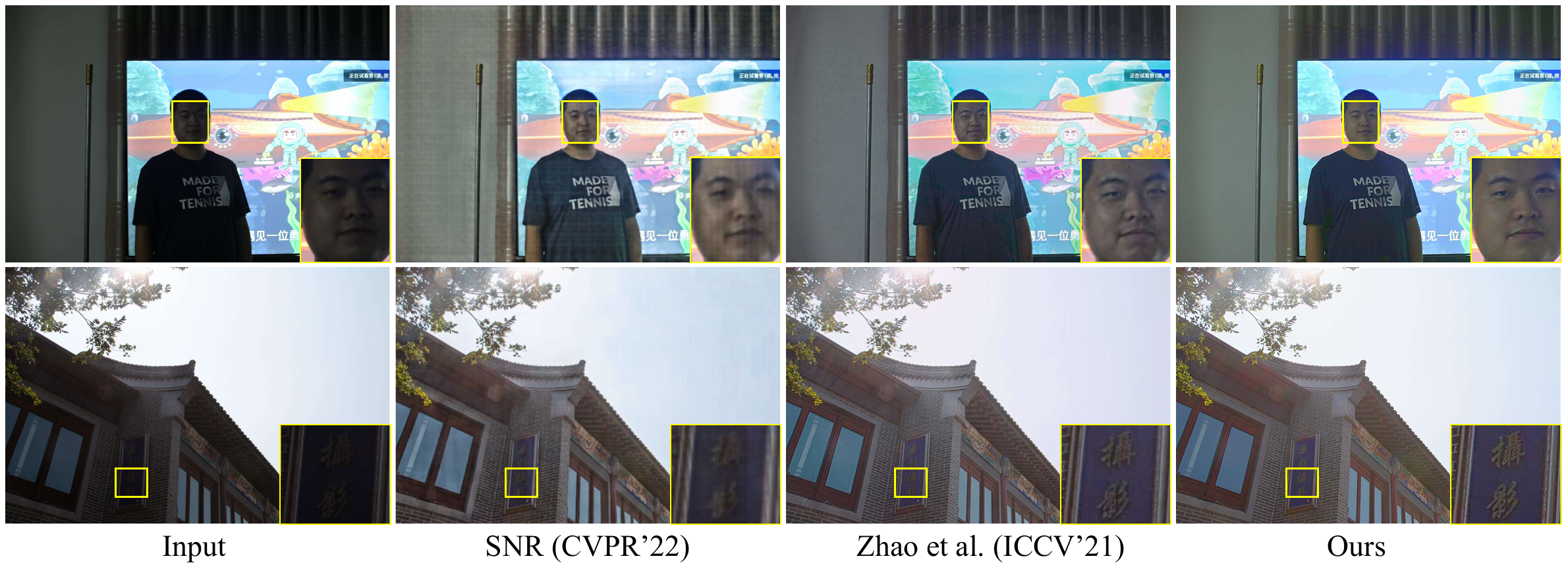}
    \end{center}
    \vspace{-1em}
    \caption{Visual comparisons between our method and the retrained supervised methods on the BAID test dataset.}
    \label{fig:comp_retrain_supervised}
\end{figure*}
\begin{figure*}[t]
\begin{center}
          \includegraphics[width=\linewidth]{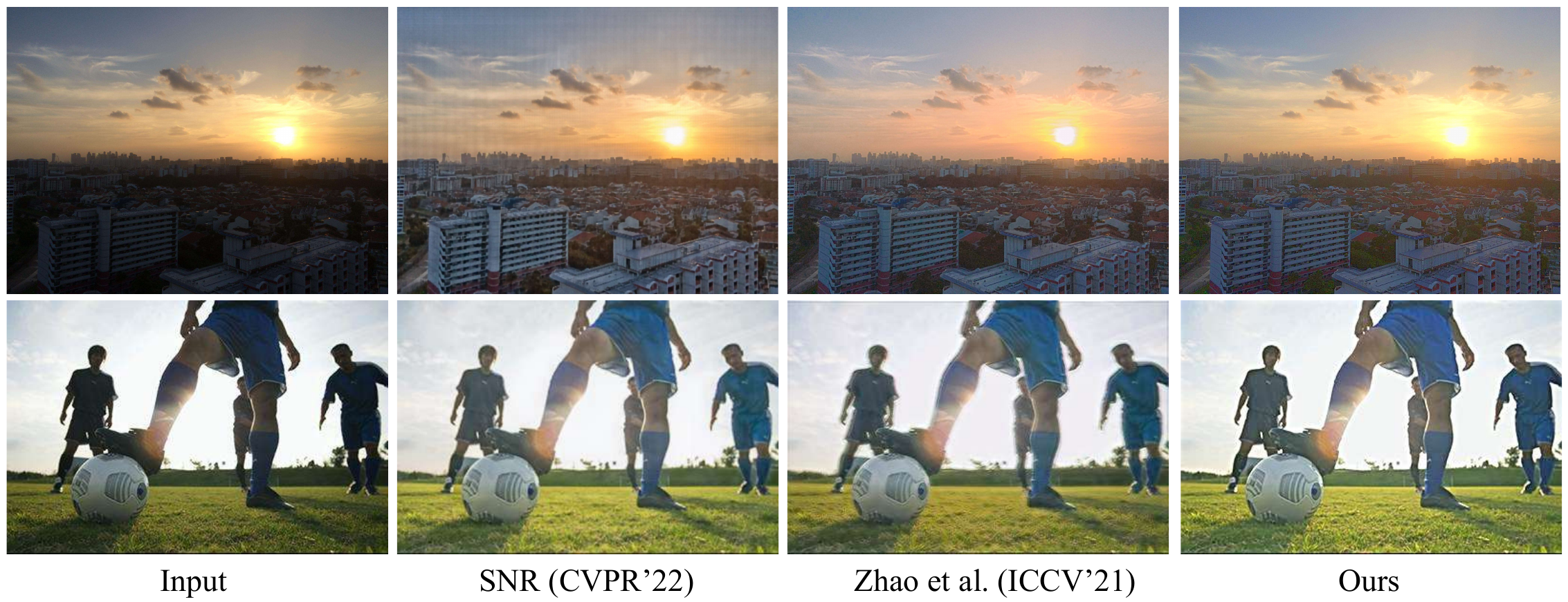}
    \end{center}
    \vspace{-1em}
    \caption{Visual comparisons between our method and the retrained supervised methods on the Backlit300 dataset.}
    \label{fig:comp_retrain_supervised_extr}
    \vspace{-0.8em}
\end{figure*}

\section{Extra Training Details}
\label{train}
\subsection{About $m_1$ settings.}
Theoretically, $m_1$ in equation (8) should be set to $0$.
However, based on our experiments, it is difficult to learn the prompt pairs under such a setting as the loss cannot converge well. 
Thus, we use a relaxed constraint $m_1=0.2$.

\subsection{Alteration Control} 
Apart from a fixed number of iterations in the process of prompt learning and enhancement training, we use two thresholds ($Thr_{A}$ and $Thr_{B}$) to control the alteration. 
Specifically, if the prompt learning's loss is lower than the threshold $Thr_{A}$, the learned prompts are frozen and the enhancement training will be triggered. 
If the enhancement model's training loss is lower than the threshold $Th_{B}$, the enhancement model is frozen and then infers the latest output images, and the prompt learning will be triggered. 
Otherwise, the alteration will be triggered when the iterations of the current process reach the pre-defined number (i.e. $1000$ for each stage). The threshold of the prompt learning loss ($Thr_A$) is set to $60$, and the threshold of the enhancement training loss ($Thr_B$) is set to $90$.

\section{Further Discussions}
\label{discussion}

Some compared supervised methods provide various models that were pre-trained on different datasets. For a fair comparison, our paper shows results from existing supervised methods without retraining, as our method does not require paired data. 
Some retrained unsupervised methods perform worse than their original models. This is primarily due to 1) some methods perform global enhancement, which may be disturbed by the uneven brightness in backlit images. 2) some methods require the training data of diverse brightness levels, while the size of our training data (380 unpaired sets) is limited. \textit{These results prove our method can perceive the pixel-level illuminance information and is free from the high requirement for sufficient training data.}

\section{Comparisons with Retrained Supervised Methods}
\label{comparision}

We retrain the state-of-the-art supervised methods on the same 380 randomly selected paired data. 
In Tab.~\ref{tab:quantitative_retrained_supervised_method}, our method outperforms the retrained methods in most metrics. 
As shown in Fig.~\ref{fig:comp_retrain_supervised} and Fig.~\ref{fig:comp_retrain_supervised_extr}, the visual results of supervised methods are relatively blurred and under-exposed. Our visual quality is superior.

\begin{table}[!h]
  \centering
    
  \resizebox{\linewidth}{!}{
  \begin{tabular}{c|c|c|c||c||c}
  \hline
   \multirow{2}{*}{ Methods}&\multicolumn{4}{c||}{BAID test dataset} & Backlit300 dataset\\
    \cline{2-6}
    & PSNR$\uparrow$ & SSIM$\uparrow$ &LPIPS$\downarrow$ &MUSIQ$\uparrow$ &MUSIQ $\uparrow$ \\
    \hline
    Zhao et al. (ICCV'21) &\best{22.123} &0.876 &0.183 &46.467 & 46.369\\
    SNR (CVPR'22)  &21.740 &0.800 &0.359 &25.930 & 31.522\\
    Ours &21.579 &\best{0.883} &\best{0.159} &\best{55.682} &\best{52.921}\\
    \hline            
  \end{tabular}
  }
  \caption{Quantitative comparison with the retrained supervised method.}
  \vspace{-1em}
  \label{tab:quantitative_retrained_supervised_method}

\end{table}

\section{More Visual Comparisons}
Here we compare our method with the other methods mentioned in our paper (i.e. all the supervised methods mentioned in our paper (Afifi et al.~\cite{Mahoud2021}, Zhao et al.~\cite{INN}, URetinex-Net~\cite{URetinexNet}, SNR-aware~\cite{SNR2022}) with all of their released pretrained models, and all the unsupervised methods mentioned in our paper ( ExCNet~\cite{zhang2019zero}, SCI~\cite{SCI2022}, Zero-DCE~\cite{Guo2020CVPR}, Zero-DCE++~\cite{ZeroDCE++},  EnlightenGAN~\cite{Jiang2019}, RUAS~\cite{RUAS2021}) as well as the version retrained on the same unpaired dataset) on the two mentioned test datasets.
\label{visual}
\subsection{Visual Comparisons on the BAID Test Dataset}
\begin{figure*}[h]
    \begin{center}
          \includegraphics[width=1.0\linewidth]{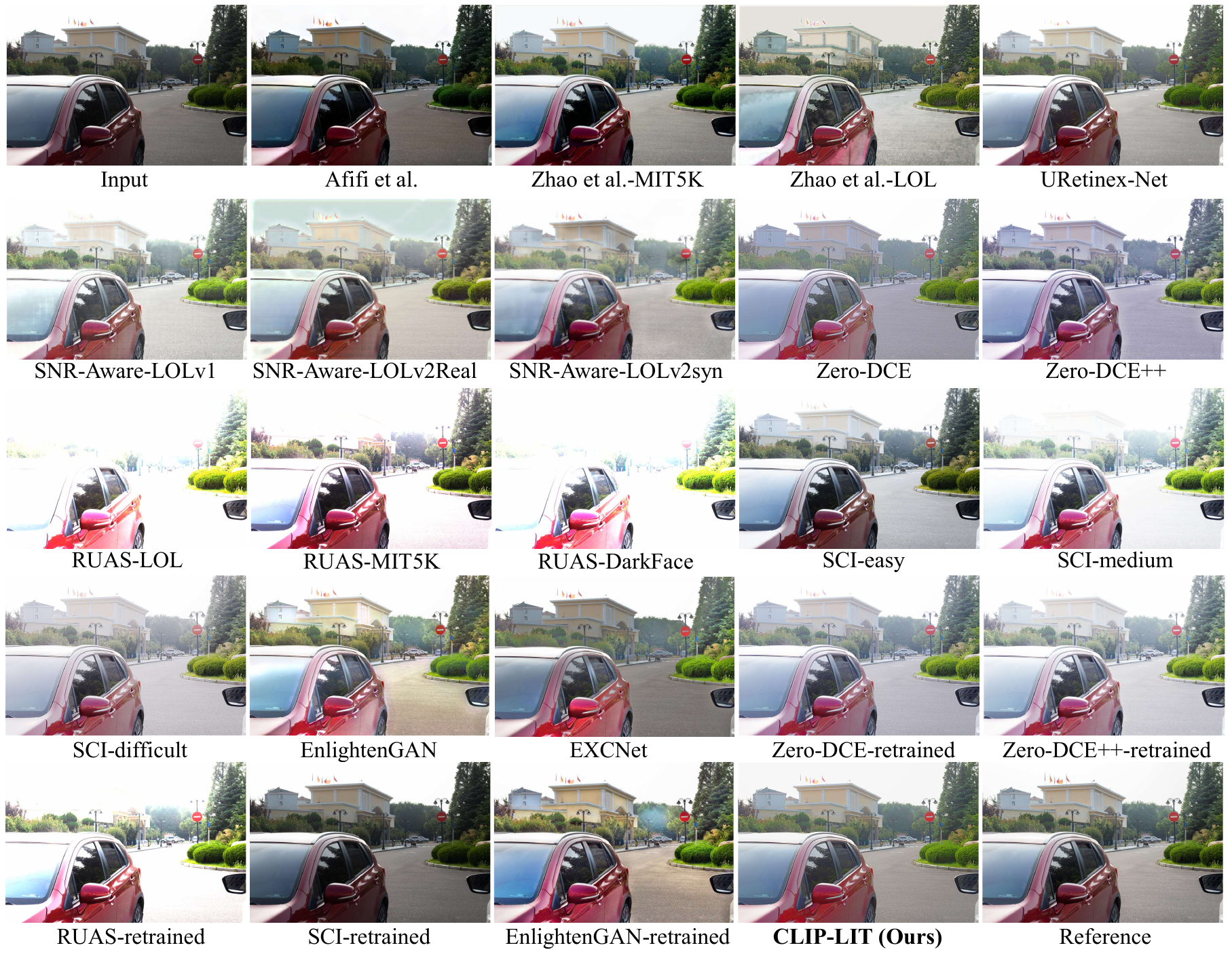}
    \end{center}
    \caption{\textbf{Complete comparisons with all methods and the reference image on the BAID test dataset.} Our CLIP-LIT effectively enhances the backlit image without causing over/under-exposure.}
    \label{fig:BAID148_25com}
\end{figure*}

\begin{figure*}[h]
    \begin{center}
          \includegraphics[width=1.0\linewidth]{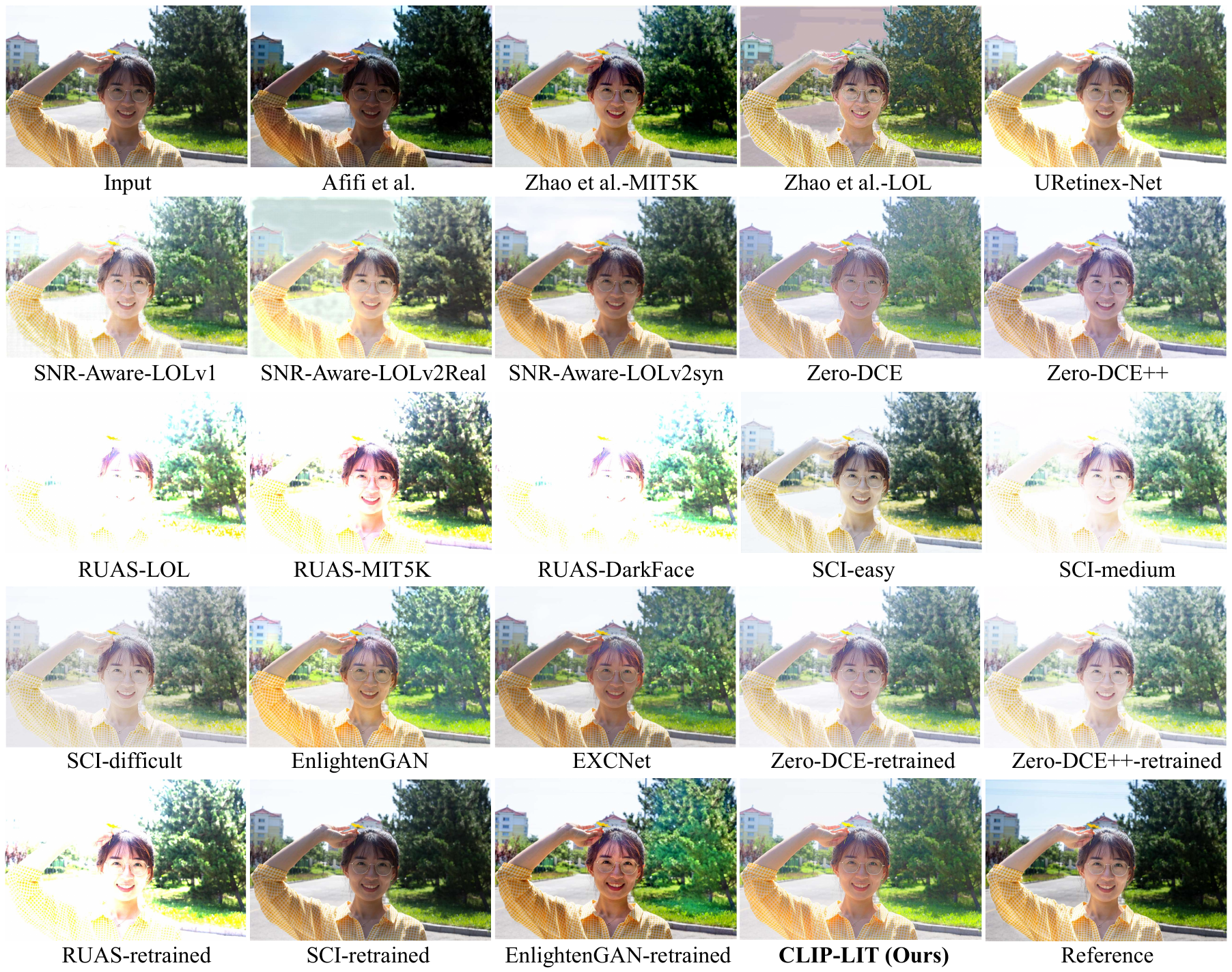}
    \end{center}
    \caption{\textbf{Complete comparisons with all methods and the reference image on the BAID test dataset.} Our CLIP-LIT produces most natural appearance than the compared methods.}
    \label{fig:BAID292_25com}
\end{figure*}

\begin{figure*}[h]
    \begin{center}
          \includegraphics[width=1.0\linewidth]{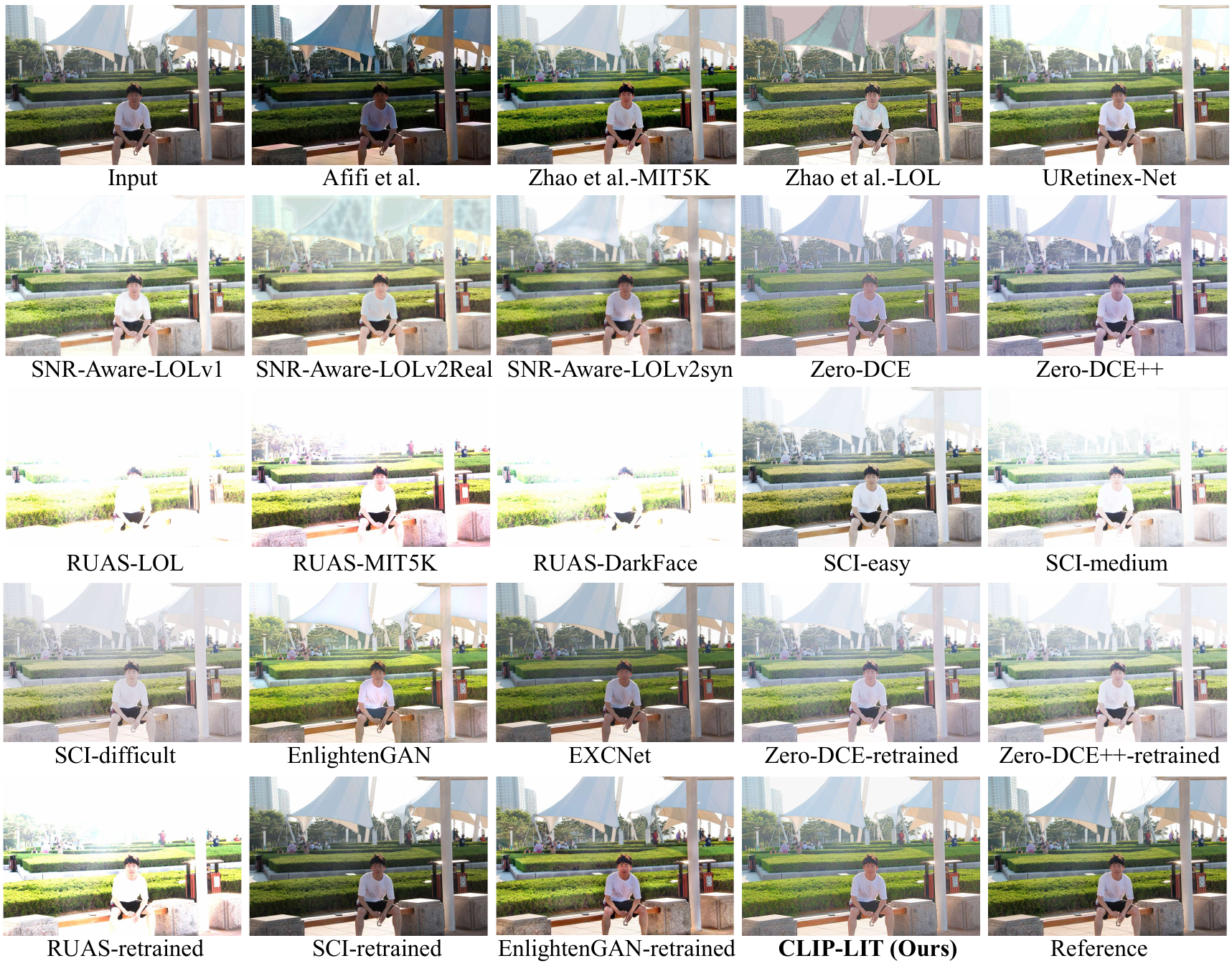}
    \end{center}
    \caption{\textbf{Complete comparisons with all methods and the reference image on the BAID test dataset.} Our CLIP-LIT restores the human face most clearly and naturally.}
    \label{fig:BAID211_25com}
\end{figure*}

\begin{figure*}[h]
    \vspace{-1.8em}
    \begin{center}
        \resizebox{15cm}{!}{
          \includegraphics[width=1.0\linewidth]{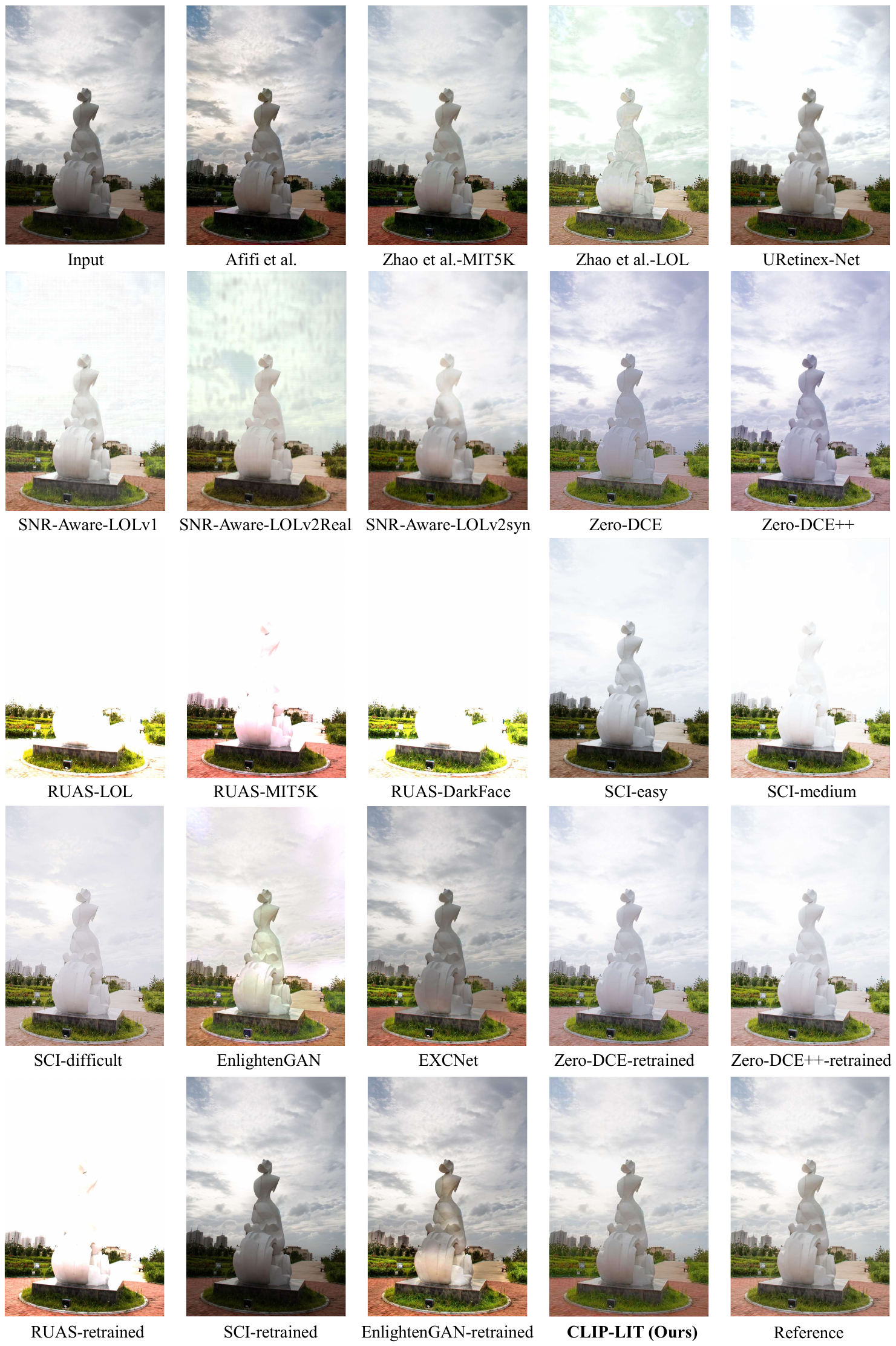}
          }
    \end{center}
    \vspace{-2em}
    \caption{\textbf{Complete comparisons with all methods and the reference image on the BAID test dataset.} Our CLIP-LIT's result is visually closest to the reference image retouched by photographers.}
    \label{fig:BAID166_25com}
    \vspace{-1em}
\end{figure*}

\begin{figure*}[h]
    \vspace{-1.8em}
    \begin{center}
        \resizebox{15cm}{!}{
          \includegraphics[width=1.0\linewidth]{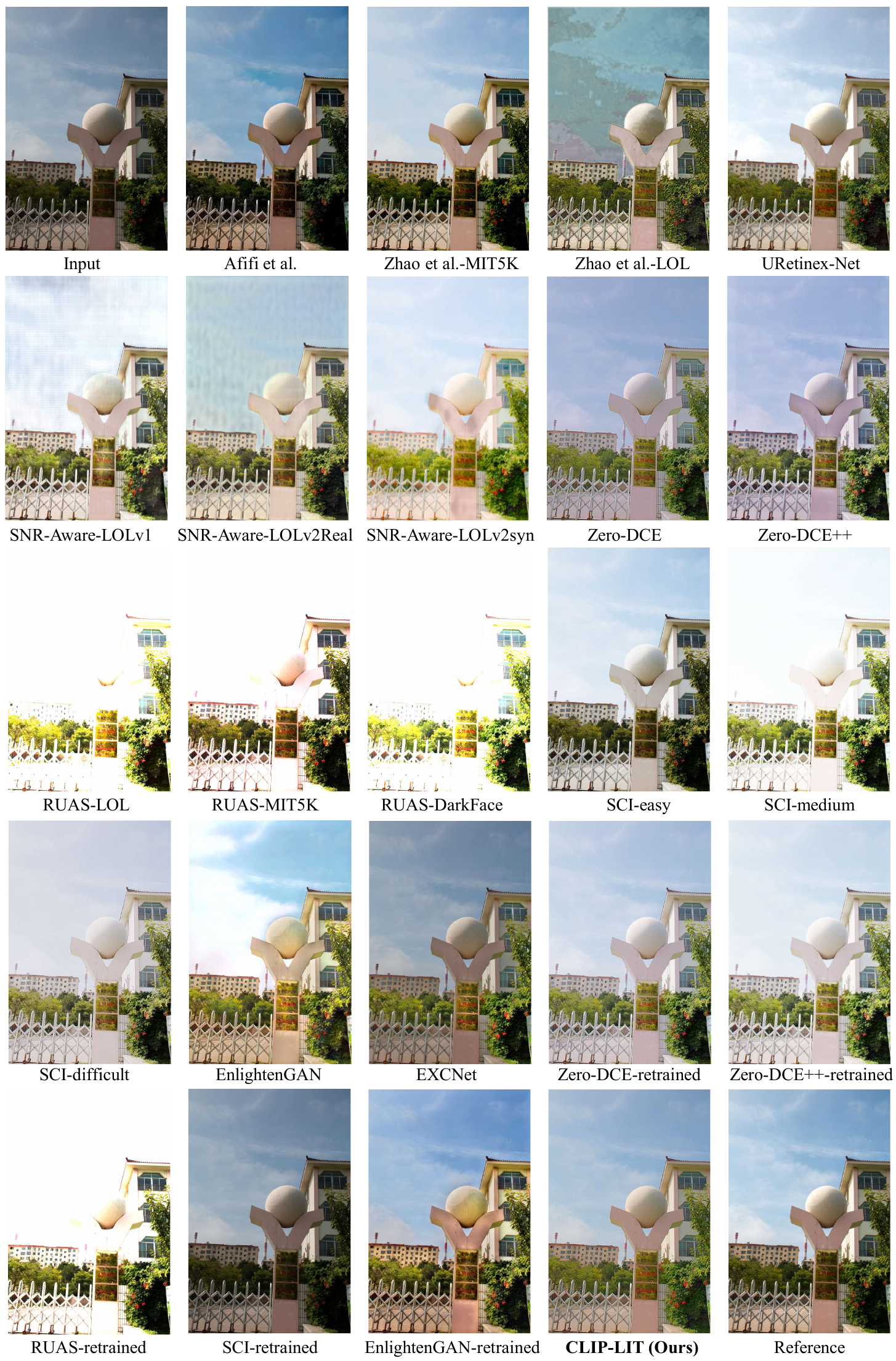}
          }
    \end{center}
    \vspace{-2em}
    \caption{\textbf{Complete comparisons with all methods and the reference image on the BAID test dataset.} Our method enlightens the dark area most naturally while preserving the color and content of the well-lit area well, that is, has the best input-output consistency in the well-lit regions.}
    \label{fig:BAID356com}
    \vspace{-1em}
\end{figure*}

\vspace{-0.5em}
Figs.~\ref{fig:BAID148_25com},~\ref{fig:BAID292_25com},~\ref{fig:BAID211_25com},~\ref{fig:BAID166_25com} and ~\ref{fig:BAID356com} show more visual comparisons between the results generated by our methods and the compared methods. Note that the BAID reference images are modified by photographers. The results show that our CLIP-LIT effectively enhances the backlit image without causing over/under-exposure and produces most natural appearance than the compared methods. 
\subsection{Visual Comparisons on The Backlit300 Dataset}
\begin{figure*}[h]
    \begin{center}
          \includegraphics[width=1.0\linewidth]{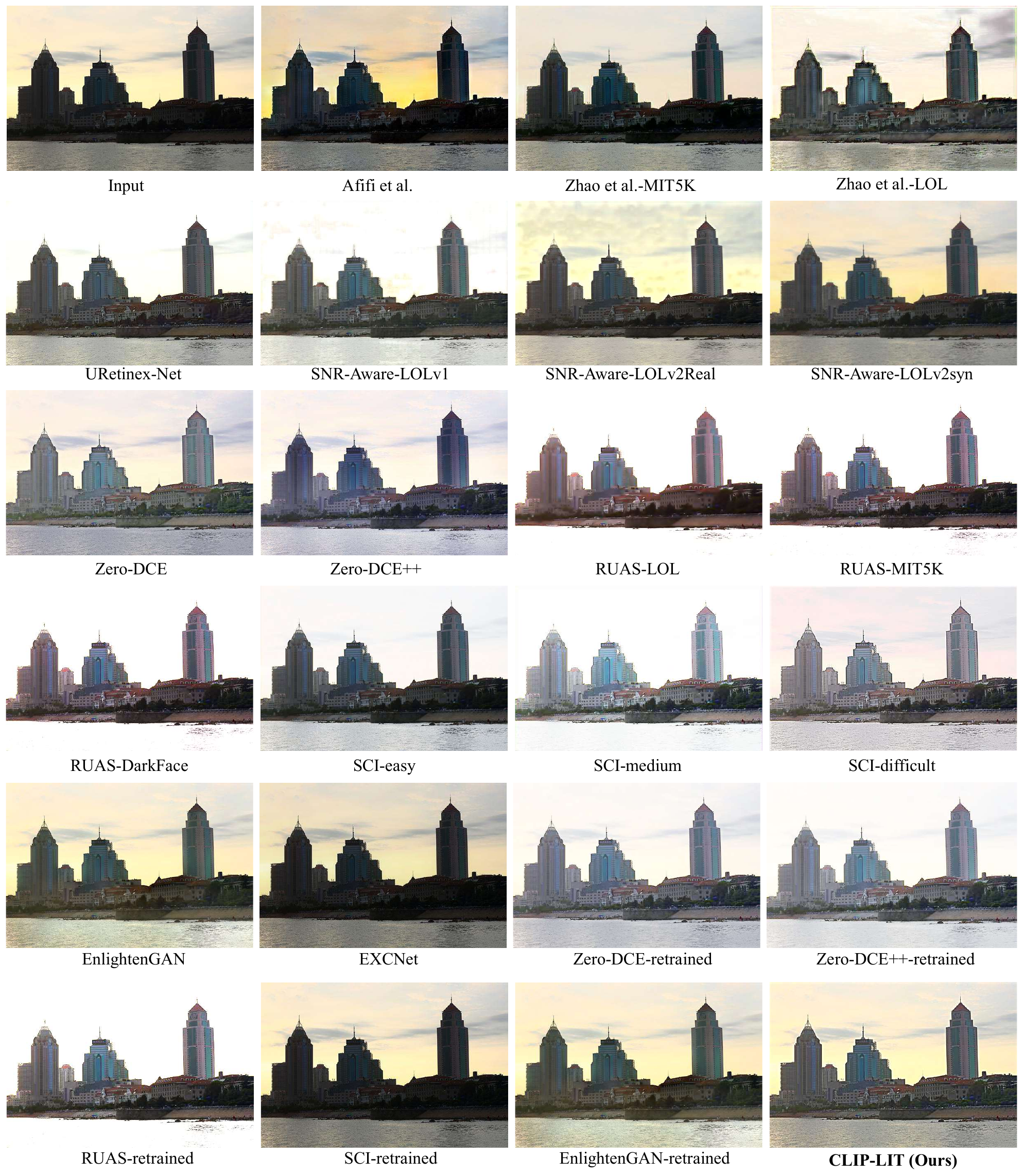}
    \end{center}
    \vspace{-2em}
    \caption{\textbf{Complete comparisons with all methods on the Backlit300 test dataset.} Our CLIP-LIT restores the color and the content of the details in the backlit area most clearly and the enhanced details have best color contrast while keeping the well-lit background remain unchanged.}
    \label{fig:extr27}
\end{figure*}

\begin{figure*}[h]
    \begin{center}
          \includegraphics[width=1.0\linewidth]{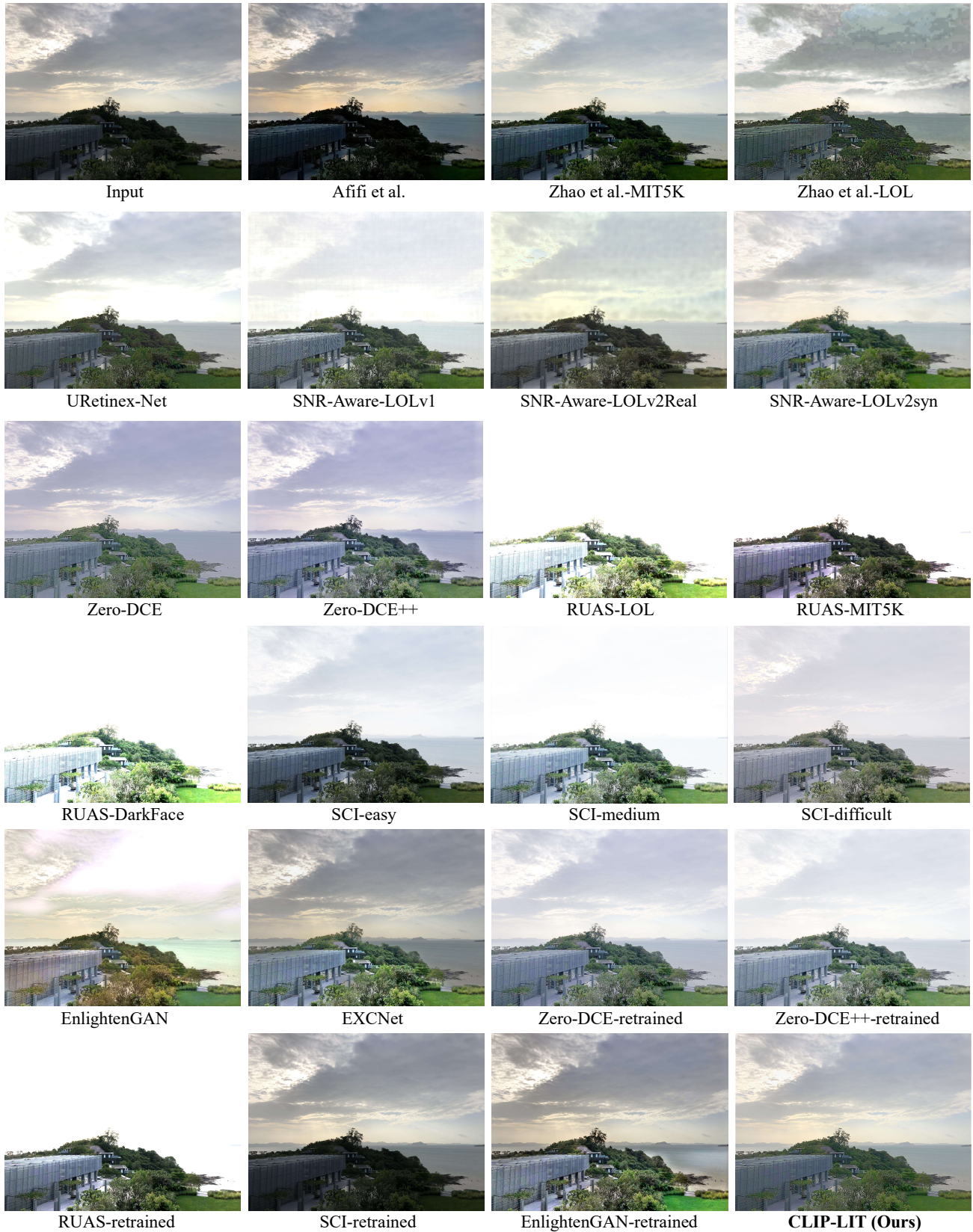}
    \end{center}
    \vspace{-1.5em}
    \caption{\textbf{Complete comparisons with all methods on the Backlit300 test dataset.} Our result is the most realistic and has input-output consistency in well-lit areas.}
    \label{fig:extr110}
\end{figure*}

\begin{figure*}[h]
    \begin{center}
          \includegraphics[width=1.0\linewidth]{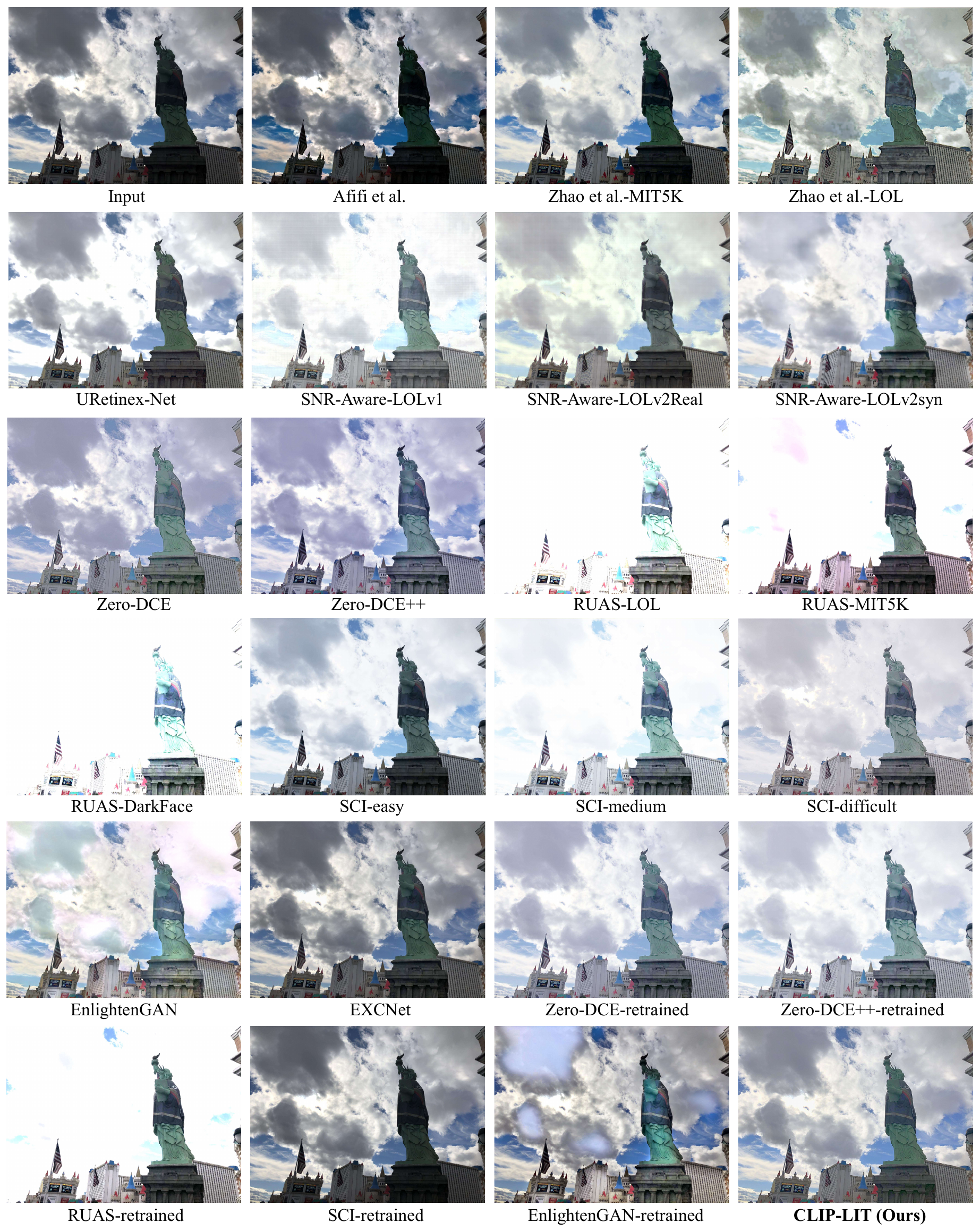}
    \end{center}
    \vspace{-2em}
    \caption{\textbf{Complete comparisons with all methods on the Backlit300 test dataset.} Our results do not contain artifacts and over-exposed regions. Moreover, Our result has the most natural color in the enlightened area and has the most pleasing contrast.}
    \label{fig:extr117}
\end{figure*}

\begin{figure*}[h]
    \begin{center}
          \includegraphics[width=1.0\linewidth]{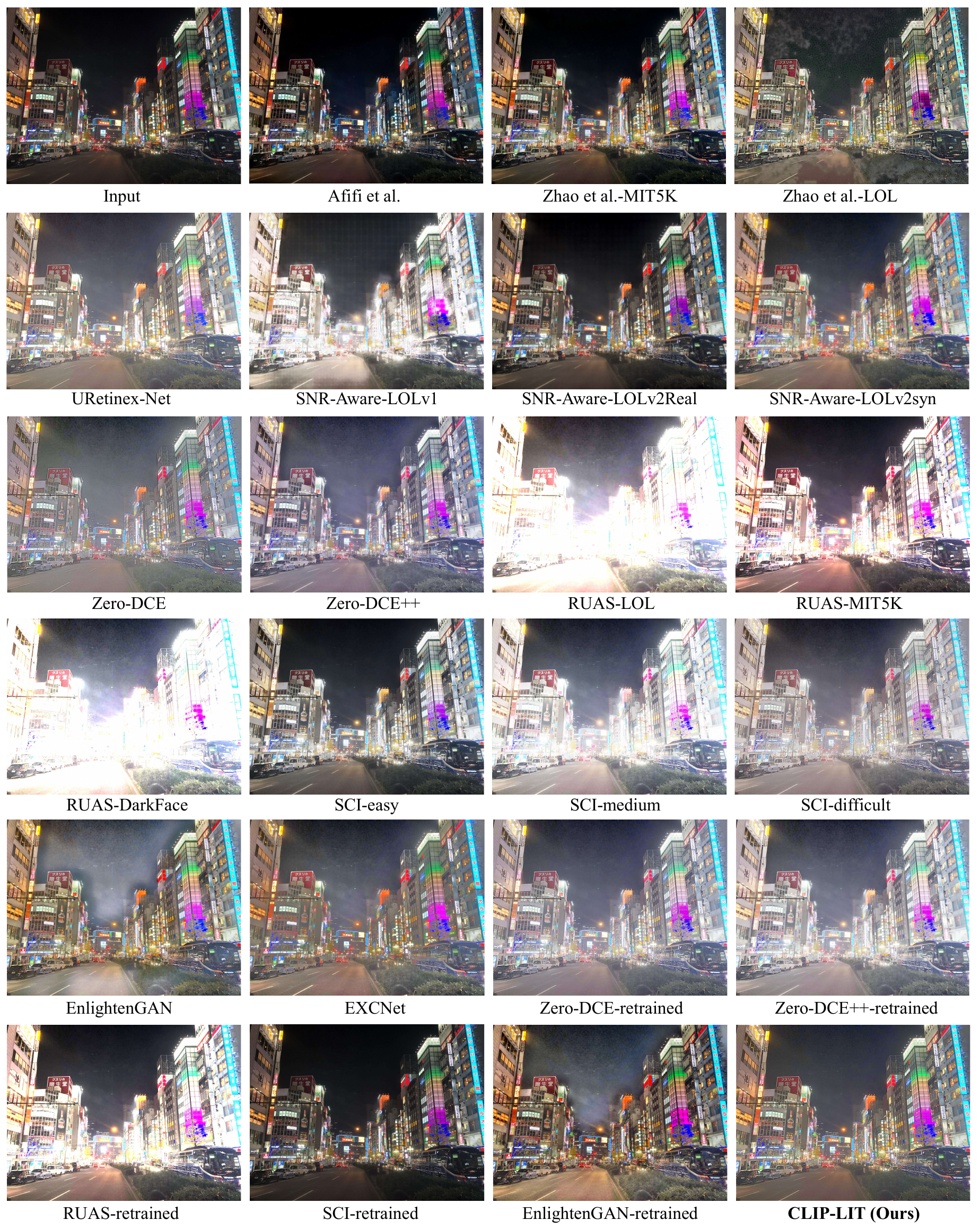}
    \end{center}
    \vspace{-2em}
    \caption{\textbf{Complete comparisons with all methods on the Backlit300 test dataset.} Our CLIP-LIT produces the most visually favorable result in the night scene as well, which enhances the backlit foreground well while not overly brightening the night sky.}
    \label{fig:extr92}
\end{figure*}

Figs.~\ref{fig:extr27},~\ref{fig:extr110},~\ref{fig:extr117}, and~\ref{fig:extr92} show more visual comparisons between the results generated by our methods and the compared methods. The results show that our method stores the color and the content of details in the backlit area most clearly and realistically, and the enhanced details have best and natural color contrast while keeping the well-lit background remain unchanged. Our CLIP-LIT yields most visually favorable result in the night scene as well.

\section{More Comparisons on More Datasets}
\label{5dataset}
\begin{figure*}[h]
    \begin{center}
          \includegraphics[width=1.0\linewidth]{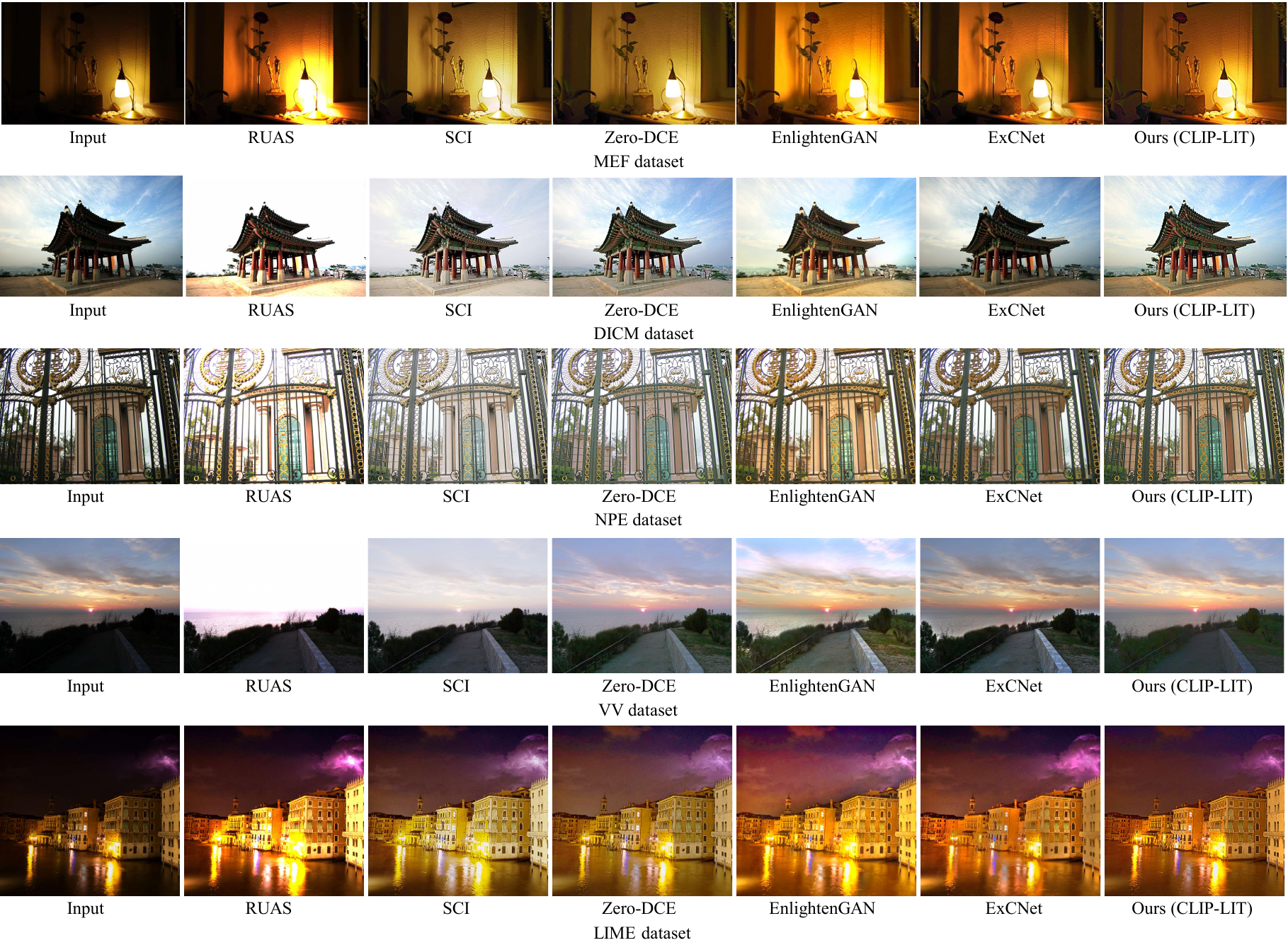}
    \end{center}
    \vspace{-1.2em}
    \caption{\textbf{More comparisons on five extra test datasets.} Our CLIP-LIT effectively enhances the unseen images without causing artifacts, over/under-exposure problems. Compared to other methods, our results have best color consistency as well.}
    \label{fig:5dataset_comp}
\end{figure*}

Here we provide both quantitative and visual comparisons with all the other unsupervised methods mentioned in our paper on five unseen extra test datasets, shown in Tab.~\ref{tab:new_dataset} and Fig.~\ref{fig:5dataset_comp}.

\begin{table}[h]
	\begin{center}
		\resizebox{1.0\linewidth}{!}{
		\begin{tabular}{c|c|c|c|c|c}
			\hline
                Dataset & DICM &LIME & MEF & NPE &VV\\
                \cline{1-6}
			Unsupervised Methods & \multicolumn{5}{c}{MUSIQ$\uparrow$}   \\ 
 			\hline
			Input &58.325 &60.092 &57.537 &59.213  &43.887   \\
 	
			\hline
                ExCNet~\cite{zhang2019zero}  &59.688 &61.590 &64.758 &62.563 &39.883\\
			Zero-DCE~\cite{Guo2020CVPR}    &59.051 &61.937 &64.923 &62.796 &38.950 \\ 
			Zero-DCE++~\cite{ZeroDCE++}   &57.794 &61.632 &63.512 &60.829 &36.379\\ 
			RUAS-LOL~\cite{RUAS2021}  &49.433 &57.315 &55.576 &50.185 &29.901 \\ 
			RUAS-MIT5K~\cite{RUAS2021}   &54.729 &60.657 &61.556 &56.785 &36.625 \\ 
			RUAS-DarkFace~\cite{RUAS2021}  &48.557 &56.216 &59.246 & 50.893 &28.668 \\ 
			SCI-easy~\cite{SCI2022} &58.684 &61.955 &63.058 &61.118 &41.423 \\ 
			SCI-medium~\cite{SCI2022}  &53.596 &60.787 &62.531 &56.726 &32.998 \\ 
			SCI-diffucult~\cite{SCI2022}  &60.309 &62.827 &65.348 &61.965 &39.308 \\ 

			EnlightenGAN~\cite{Jiang2019} &57.024 &60.775 &62.978 &59.300 &32.337 \\ 

			CLIP-LIT (Ours) &\best{61.136}  &\best{63.761} &\best{66.359} &\best{63.798} &\best{42.848}\\ 
			\hline
		\end{tabular}
		}
	\end{center}
	\vspace{-1em}
 \caption{Quantitative comparisons on the \textbf{five extra test datasets}.}
 \label{tab:new_dataset}
 \vspace{-1em}
\end{table}
The results indicate the performance of our method still reaches SOTA on these five unseen test datasets. And as shown in Fig.~\ref{fig:5dataset_comp}, compared to other unsupervised methods, our CLIP-LIT yields most natural results, without causing artifacts and over/under-exposure problems.

\section{Discussions of failure cases and potential extensions}
\label{extension}
Our method may fail in processing extreme cases such as the information missing in over-/under-exposed regions due to the limited bit depth
of sRGB image. Extending our method to HDR data may be a potential solution to such cases. 

Our method cannot denoise since our training data doesn't contain noise. Adding noise argumentation may address this problem.

We believe our method can also be a possible way to help other image restoration tasks, such as low light enhancement, deraining, etc.

\section{Additional Illustration}
Our model is lightweight and can handle a 4K image within 0.5 seconds using a single NVIDIA GeForce RTX 3090 GPU. We provide some 4K video samples enhanced by our CLIP-LIT in our video demo. Due to size limits, we resize the outputs to 2K to show in the demo.




\clearpage



\end{document}